\documentclass[10pt,twocolumn,letterpaper]{article}

\usepackage[pagenumbers]{cvpr} 

\usepackage{graphicx}
\usepackage{amsmath}
\usepackage{amssymb}
\usepackage{booktabs}

\usepackage[pagebackref,breaklinks,colorlinks]{hyperref}

\usepackage[capitalize]{cleveref}
\crefname{section}{Sec.}{Secs.}
\Crefname{section}{Section}{Sections}
\Crefname{table}{Table}{Tables}
\crefname{table}{Tab.}{Tabs.}

\def\cvprPaperID{*****} 
\def\confName{CVPR}
\def\confYear{2022}

\begin{document}

\title{Point-Based Radiance Fields \\
for Controllable Human Motion Synthesis
}

\author{Haitao Yu\\
D-MATH ETHz\\
{\tt\small haityu@student.ethz.ch}
\and
Deheng Zhang\\
D-INFK ETHz\\
{\tt\small dezhang@student.ethz.ch}
\and
Peiyuan Xie\\
D-MATH ETHz\\
{\tt\small peixie@student.ethz.ch}
\and
Tianyi Zhang\\
D-MAVT ETHz\\
{\tt\small zhangtia@student.ethz.ch}
}
\maketitle

\begin{abstract}
   This paper proposes a novel controllable human motion synthesis method for fine-level deformation based on static point-based radiance fields. Although previous editable neural radiance field methods can generate impressive results on novel-view synthesis and allow naive deformation, few algorithms can achieve complex 3D human editing such as forward kinematics.  Some algorithms such as \cite{wang2022neural} can edit the human using the SMPL model, but cannot be readily generalized to other 3D characters without a prior SMPL model. Our method exploits the explicit point cloud to train the static 3D scene and apply the deformation by encoding the point cloud translation using a deformation MLP. To make sure the rendering result is consistent with the canonical space training, we estimate the local rotation using SVD and interpolate the per-point rotation to the query view direction of the pre-trained radiance field. Extensive experiments show that our approach can significantly outperform the state-of-the-art on fine-level complex deformation which can be generalized to other 3D characters besides humans. To better observe our performance, we highly recommend readers see the \href{https://youtu.be/3iZ_89IwZUU}{videos} which demonstrate the result. We also provide our code in the following link: \href{https://github.com/dehezhang2/Point_Based_NeRF_Editing}{https://github.com/dehezhang2/Point\_Based\_NeRF\_Editing}. 
\end{abstract}

\section{Introduction}
\label{sec:intro}

Synthesizing motions of digital humans and other 3D characters in virtual scenes is of great significance to many computer vision and graphics applications, including visual effects, video games, and telepresence. Many of these applications would desire easy control over the poses of the characters while achieving photo-realistic rendering in a reasonable time and maintaining the environment in the original scene. Recent progress in editable scene reconstruction has displayed promising potential to alleviate expensive manual invention and corrections in this process, but it still remains quite challenging to produce results of those qualities at the same time.

Neural radiance field (NeRF) \cite{mildenhall2020nerf} and its extensions \cite{barron2021mip,verbin2022ref} have demonstrated the capability of modeling real scenes from multi-view images with implicit neural representations. While having shown great success in synthesizing impressive novel views, most of their applications are limited to static scenes and thus do not allow user control. On the other hand, point-based methods \cite{su2023npc, xu2022point} for building NeRF models proved to have better rendering performance, compared with their purely implicit or volumetric counterparts. Moreover, point clouds as primitives provide the possibility for users to directly manipulate and deform the human characters in the scene to their needs.

In this paper, we extended the point-based radiance fields with the ability to synthesize user-controllable human motion. In point-based radiance fields, the view-dependent color of each point is modeled with the absolute viewing direction. Based on this observation, we propose a method to bend the ray direction back to the original scene when rendering the deformed scenes, which enables us to produce more visually plausible and physically convincing views of scenes with novel human motions.

Specifically, our method builds a radiance field from static scenes based on PointNeRF. In the point-based representation, each point is associated with a neural feature which is defined as a function of viewing direction and encodes the local scene content. Deforming the pose can be naturally and easily achieved by directly manipulating the points of interest and moving them to the new positions, which have direct benefits over existing mesh-based NeRF editing methods in terms of efficiency. However, the point cloud generated from PointNeRF is dense and noisy for the deformation. Furthermore, user editing is not intuitive to make complex deformation. In order to efficiently edit the point cloud, we use a neural network to model the deformations of the point cloud and incorporate it into our pipeline. It has been shown that learning deformation networks on point sets avoids restricting the space of deformations to linear blend skinning functions and the limitations on human characters, while still remaining simple and computationally attractive. It also allows users to control the motion through a set of key points (user-defined or SMPL), ridding them of tediously manipulating every single point. Moreover, we demonstrated the possibility of a point-based radiance field to render scenes with both static backgrounds and deformed characters. 

While point-based deformation is quite intuitive and straightforward, direct rendering without further processing could lead to inconsistent results if the deformation consists of large rotation, due to the change of viewing direction of each point. A number of approaches have been proposed to tackle this problem, by twisting the ray back in the canonical space or adding extra information into the feature at training. However, these methods introduce extra complexity into NeRF models and strongly limit their application in practice. Instead, our method tries to add a ray-bending module to the current point cloud-guided NeRF model. Our key idea is the following: with the deformed and original point clouds, we could estimate the local rotation of each point during the deformation. This information then enables us to redirect the querying ray in the right direction, ensuring the consistency of the learned implicit representation in the deformed space with that in the canonical space. 

To summarize, our main contributions are as follows. We first introduce ray-bending to the current point-based radiance field, together with a deformation field represented by a neural network, which can be used to synthesize controllable and visually consistent human motion. The advantages of our proposed approach are demonstrated by comparing it to the methods without considering the change of view dependence at rendering after deformation. Second, we showcase the ability of our method to be applied directly to nonhuman data, and the possibility to render the deformed human characters with the background. Last but not least, we provide a whole pipeline for users to generate datasets and test out algorithm.


\section{Related Work}
\label{sec:relatedwork}

\subsection{Neural Radiance Fields}
Previous work NeRF \cite{mildenhall2021nerf} proposed a radiance field-based representation of the 3D scene through an implicit neural network. The method is trained using multi-view 2D images for which the camera pose is provided and uses two MLPs to predict the density $\sigma$ and radiance $r$ of a given sample on the ray. Then the volumetric rendering equation is applied to render the image for different views. Follow-up works modify the structure to either improve the rendering quality or speed up the training of NeRF. For example, Mip-NeRF \cite{barron2021mip} applies cone sampling instead of ray sampling to deal with the aliasing issue, Ref-NeRF \cite{verbin2022ref} uses the reflected direction of the camera ray as the input of MLP to improve the rendering quality for specular materials, \cite{fridovich2022plenoxels, chen2022tensorf, muller2022instant} modify the data structure of the radiance field to improve the training speed of NeRF.  However, the implicit representation limits fine-grained scene editing based on mesh deformation. Other works \cite{park2021hypernerf,pumarola2020d,park2021nerfies,peng2021animatable,wang2022arah} can learn the dynamic scene given the dynamic multi-view ground truth for supervision. Nevertheless, these methods cannot readily apply complex deformation defined by users to the scene. In our work, we utilize a point-cloud data structure, which decouples the feature 3D position from the implicit representation, making fine-grained scene editing plausible.

\subsection{Point-Based Representation}
In order to decouple the 3D position from the implicit MLP, recent works \cite{xu2022point, kerbl20233d,zhang2022differentiable,chen2023neuraleditor} introduced point cloud as the data structure to encode the position-dependent scene information. For example, \cite{xu2022point, chen2023neuraleditor} propose a pipeline, which first initializes a point cloud with point features using multi-view stereo methods, then optimizes the MLP and features during volumetric rendering. During the optimization, the point cloud is further optimized by a pruning and growth strategy. While \cite{kerbl20233d, zhang2022differentiable} use differentiable rasterization to render the point clouds and optimize the corresponding point features. However, these methods lack fine-grained controllability for the users to precisely edit the point cloud to achieve complex deformation. In our approach, we apply a keypoint-based deformation model to the point cloud, together with ray-bending to consistently render the deformed point cloud using the pre-trained point radiance field. 
\subsection{Scene Editing via NeRFs}
To improve the diversity of the scene rendering, many follow-up works of NeRF \cite{mildenhall2021nerf} explore different aspects of scene editing. For instance, CLIPNeRF \cite{wang2022clip} deforms the position and direction of the query using CLIP  \cite{radford2021learning} encoder and optimizes the radiance via CLIP-based loss function, whereas \cite{kobayashi2022decomposing} distills the 2D semantic feature from Lseg \cite{li2022language} to train 3D semantic feature using volumetric rendering, enabling editing including colorization, translation, deletion, and text-driven editing together with CLIPNeRF \cite{wang2022clip}. \cite{zhang2022arf, wang2022nerf} defines a 2D-based style loss function to achieve style transfer editing on the pre-trained NeRF radiance. However, these methods mainly focus on radiance editing, and geometric editing is not available.

Other works, such as \cite{lazova2023control} allow compositing objects from different scenes by sharing the radiance and density prediction network. \cite{kania2022conerf} enables dynamic facial expression control using 2D annotations. To enable more complex deformation, \cite{yuan2022nerf, garbin2022voltemorph, peng2022cagenerf, xu2022deforming, jambon2023nerfshop} use the cage or tetrahedron-based barycentric coordinates to edit the query for NeRF trained in static scenes, achieving user-defined deformation on the given scene. However, these methods do not change the intrinsic state of NeRF, and the deformation is applied by changing the query of ray marching through interpolation of cage deformation. Therefore, only coarse-level deformation can be achieved, and the user cannot apply complex deformation such as forward kinematics. Another recent work \cite{wang2022neural}, enables more complex user editing with a SMPL \cite{SMPL:2015} conditioned network. This method has better controllability, but cannot be generalized to non-human objects since SMPL prior is required. Benefiting from the explicit 3D position representation in PointNeRF \cite{xu2022point}, we propose a novel algorithm that enables fine-level deformation with key point guidance. Different from another recent work \cite{chen2023neuraleditor}, which also applies deformation based on the point cloud, our method utilizes an MLP to interpolate the deformation of the keypoint, which achieves more complex user editing that is generalizable to all kinds of objects instead of simple local deformation.
\subsection{3D Shape Deformation}
Since we directly apply the deformation to the neural point cloud, our method is highly related to 3D shape deformation. Traditional methods such as \cite{jacobson2012fast,jakab2021keypointdeformer,ju2005mean,merry2006animation,thiery2014jacobians,yifan2020neural,zhang2020proxy,igarashi2005rigid} can apply controllable deformation to the triangular meshes. However, these methods cannot be directly generalized to the point cloud without edge defined. A recent work Dynamic Point Field (DPF) \cite{prokudin2023dynamic}  shows extraordinary deformation results on point clouds with key points as supervision. Similar to DPF, our work uses an MLP to predict the translation between the deformed space point cloud and the canonical space point cloud. However, to consistently render the deformed space, we borrow the idea from ARAP \cite{igarashi2005rigid} to estimate location rotation and apply ray bending to the query. Our experiment shows that the deformation and rendering quality is superior for fine-level user editing. 

\section{Method}
\begin{figure*}
  \centering
\includegraphics[width=\textwidth]{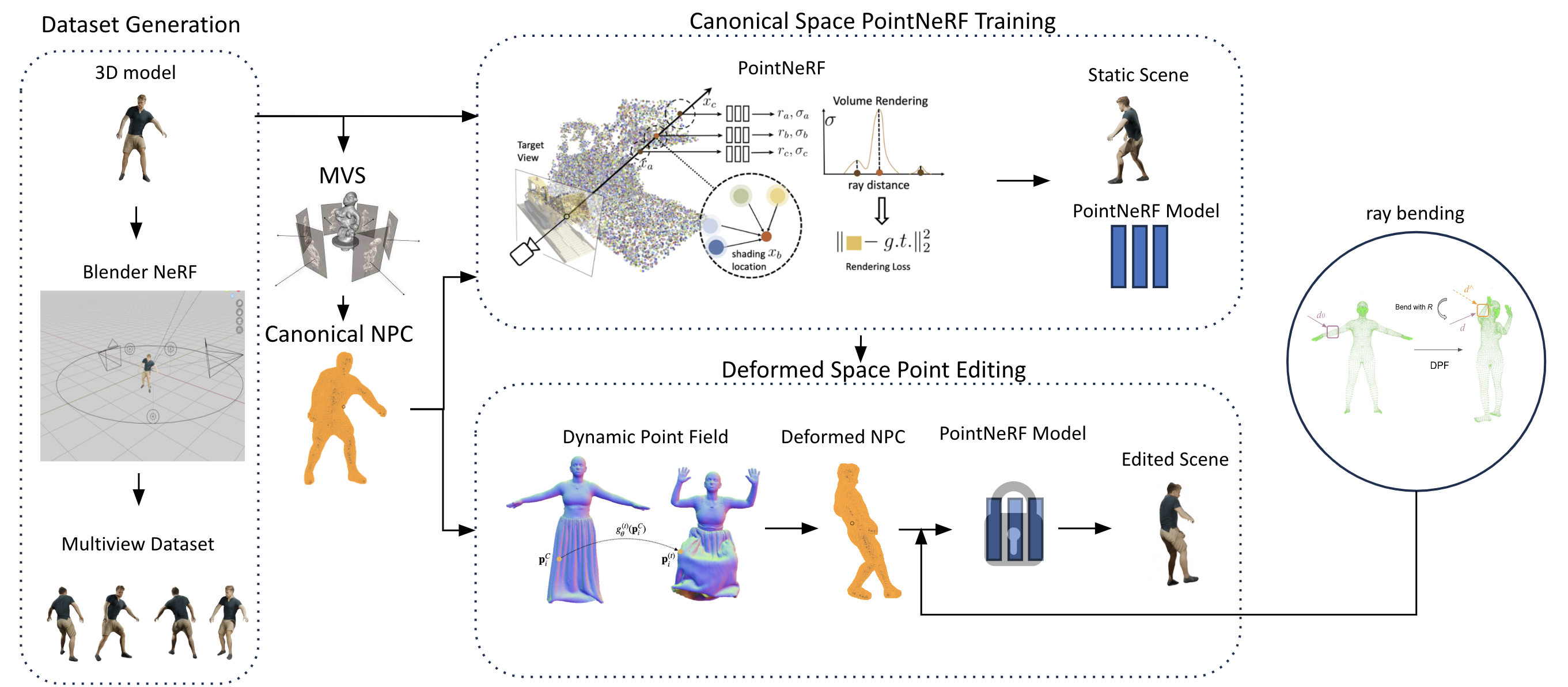}
  \caption{\textbf{Our pipeline.} (1) We first use Blender NeRF \cite{Raafat_BlenderNeRF_2023} to generate the multiview dataset. (2) We generate the initial canonical space neural point cloud using a multi-view stereo algorithm and train the PointNeRF \cite{xu2022point} for the static scene. (3) We use a deformation MLP similar to DPF \cite{prokudin2023dynamic} with key points as input to edit the canonical point cloud. Then we apply ray bending to the deformed point cloud and render the deformed scene using the pre-trained PointNeRF model.}
  \label{fig:pipeline}
\end{figure*}

\subsection{Overview}
In this project, we use neural points as primitives, which explicitly represent the scene geometry and implicitly store the radiance information. Generally, this representation is capable of bridging high-fidelity scene reconstruction and complex deformation models together, in order to achieve controllable motion synthesis and photo-realistic rendering at the same time. Our pipeline is shown in Fig. \ref{fig:pipeline}, we first use Blender NeRF \cite{Raafat_BlenderNeRF_2023} to generate the multiview dataset. Then, we generate the initial canonical space neural point cloud using a multi-view stereo algorithm and train the PointNeRF \cite{xu2022point} for the static scene. During deformation, we use a deformation MLP similar to DPF \cite{prokudin2023dynamic} with key points as input to edit the canonical point cloud. Then we apply ray bending to the deformed point cloud and render the deformed scene using the pre-trained PointNeRF model.

\subsection{Point-based Reconstruction}

\textbf{Initialization and optimization of the point cloud.}
In order to use neural points to represent the scene geometry, we first need to generate 3D point locations and the confidence of each point. For this, we leverage deep MVS methods using deep 3D CNNs. Consistent with PointNeRF, as for learning the radiance information of the point cloud, we use another 2D CNN to extract feature maps from images, which are then aligned with the generated points and are used to describe per-point features. After the initialization, point growing and pruning procedures will be performed to eliminate outliers and holes, ensuring the rendering quality.

\textbf{Incorporating background.}
To include the background in the scenes, we tried combining our current pipeline with different MVS models, such as MVSNet \cite{yao2018mvsnet}, COLMAP \cite{schonberger2016structure}, and TransMVS \cite{ding2022transmvsnet}, in order to achieve better initialization. However, these models cannot deal with regions with similar colors very well, which is a common case in the real world. And the point growing and pruning procedures mentioned above fail to compensate for this. So instead, we first represent the background with points extracted from the mesh, keep track of the points belonging to humans and background separately, and in the end combine them during final rendering.
\subsection{Point-based Rendering}
To achieve novel-view synthesis for canonical space, we use the same approach as PointNeRF\cite{xu2022point}. We denote the neural point cloud by $P = \{(x_i, f_i, \gamma_i) |i=1,...,N\}$, where each point $i$ is located at $x_i$ with a neural feature $f_i$ and a scale confidence value $\gamma_i \in [0,1]$. For the static scene, our density and radiance prediction is the same as PointNeRF:
\begin{align*} 
      (\sigma, r) = \text{Point-NeRF}(x,v, x_1, f_1, \gamma_1,...,x_k, f_k, \gamma_k),
\end{align*}
where $x$ represents the 3D location of the query point in canonical space, and $v$ represents the canonical view direction during rendering. However, after deformation, the view direction $\hat{v}$ should be transformed to canonical space since we train the Point-NeRF in canonical space. Therefore, the prediction in deformed space becomes:
\begin{align*} 
    &v = \text{Ray-Bending}(\hat{v})\\
    &(\sigma, r) = \text{Point-NeRF}(\hat{x},v, x_1, f_1, \gamma_1,...,x_k, f_k, \gamma_k),
\end{align*}
where $\hat{x}$ represents the 3D location of the query point in deformed space, and $\hat{v}$ represents the deformed view direction during rendering. Then, the predicted radiance and density are rendered by the volume rendering equation. 
\subsection{Point-based deformation}

Unlike other space-based deformation models adopted by common deformable scene representation, this paper utilizes the idea of surface-based deformation by taking advantage of the explicit geometry reconstruction of the scene with the neural points mentioned above. In this work, we use a similar approach to Dynamic Point Field (DPF) \cite{prokudin2023dynamic} as our deformation model. DPF is capable of efficiently modeling highly non-linear, expressive human motions under the setting of point-based representation.

As to the usage of DPF, we first reasonably extract sparse key points set $\{\hat{\textbf{x}}_i\} \subseteq \hat{\textbf{X}}$ with size  $n$  from the canonical space where static NPC (neural point cloud) belongs to. Another input of DPF is the provided $T$ target pose keypoints frames ${\{\textbf{x}^{(t)}_i\} \subseteq \textbf{X}^{(t)}, t = 1,\ldots, T }$ which have fine correspondence with the source key points. The job of DPF here is to output a series of compact neural networks $\{g_{\theta}^{(t)}, t = 1,\ldots, T \}$ that encodes the translation from the canonical space $\hat{\textbf{X}}$ to the deformed space $\textbf{X}^{(t)}$ encoding the human motion. 
\begin{align*} 
      &g_{\theta}^{(t)} : \mathbb{R}^{ 3} \rightarrow \mathbb{R}^{ 3}\\
      &\textbf{x}^{(t)}_i = \hat{\textbf{x}}_i + g_{\theta}^{(t)}(\hat{\textbf{x}}_i)
\end{align*}

We believe that the $ g_{\theta}^{(t)}$ approximators will be sufficiently generic to propagate the deformation to canonical space containing whole dense NPC. Consequently, static NPC will be mapped to the deformed space accordingly, with their attached radiance information untouched, in that this process is completely decoupled from the PointNeRF training. The ensuing evaluation/rendering of the PointNeRF will march the rays to the position-updated NPC, which explicitly encodes the human motion.

\subsection{Ray bending}

Ray bending is a common rendering paradigm for editable NeRF and its analogs due to their implicit nature, which means the scene objects are not actually deformed but another way around. In specific, the ray cast from the camera is bent along the volumetric marching path by the absolute transformation induced by the deformation, in order to query the canonical space where the actual NeRF residents. Here, transformation stands for translation and rotation, which correspond to the two inputs of NeRF rendering: spatial coordinate and view direction.

In this work, coping with the translation part of the ray bending is not necessary anymore, since NPC are explicitly deformed with DPF. The rotational part of ray bending still remains to be addressed, in the case of view-dependent information. Generally, we will estimate the relative rotation at each queried NPC first and apply it to the corresponding view direction to realize the consistent querying in canonical space.

\textbf{Relative rotation estimating}: The relative rotation at each NPC is estimated in a geometry manner, by extracting a small local region of the interested surface in the canonical space and finding its corresponding part in the deformed space. Since the point cloud does not have any adjacency topology, we use k-nearest-neighbor as our interested cluster instead. We build a KD-tree structure similar to \cite{chen2023neuraleditor} for the deformed point cloud for each frame. And for each point $\textbf{c}_{\hat{x}}$ in the deformed point cloud, we query the KD-tree to extract $k$-nearest-neighbor $\hat{X}$ in the deformed space. We denote the corresponding point and neighbors in canonical space as $\textbf{c}_{x}$ and $X$. Then SVD can be used upon two clusters to obtain the estimated rotation matrix, which is applied to the deformed space view direction $\hat{\textbf{v}}$ to obtain canonical space view direction $\textbf{v}$ as follows:

\begin{align*}
     U, E, V &= \text{SVD} ((X - \textbf{c}_{x})^T(\hat{X} - \textbf{c}_{\hat{x}})),\\
     R &= UV^T, \text{ s.t.}\\
     \textbf{v} &= R\hat{\textbf{v}}, \quad \text{where clustering }\hat{X},X \in \mathbb{R}^{k \times 3}
\end{align*}

\textbf{Rotation Interpolation for world query}: In PointNeRF rendering setting, the volumetric ray marching will not directly query the specific NPC at each sample point but use the KNN-based aggregated feature. Similarly, We interpolate the relative rotation for each sample in ray marching $R$ using $k$-nearest-neighbor in NPC as well. We use inverse-distance weighted NLerp between several different rotations to preserve the orthogonality of the interpolated rotation $R$ in $\mathcal{SO}(3)$. Let $R_i, i \in \{1,\ldots,k\}$ denote the $k$ queried rotation matrices and $f$ be the mapping from rotation matrix to the quaternion, rotation interpolation for world query can be formulated as followings:

\begin{align*}
     q_i &= f(R_i),\\
     \quad q &= {\sum_{i=1}^k w_i q_i \over ||\sum_{i=1}^k w_i q_i||},\\
     R &= f^{-1}(q), \\
     \text{where } &w_i = \frac{d^{-1}_i}{\sum_{i=1}^k {d^{-1}_i}}
\end{align*}

The validity of ray bending will be demonstrated in the ablation study section.

\begin{figure*}[h]
    \centering
        \begin{tabular}{p{12mm}p{18mm}|p{12mm}p{12mm}p{12mm}p{12mm}p{12mm}p{12mm}p{12mm}p{12mm}}
             \includegraphics[width=20mm]{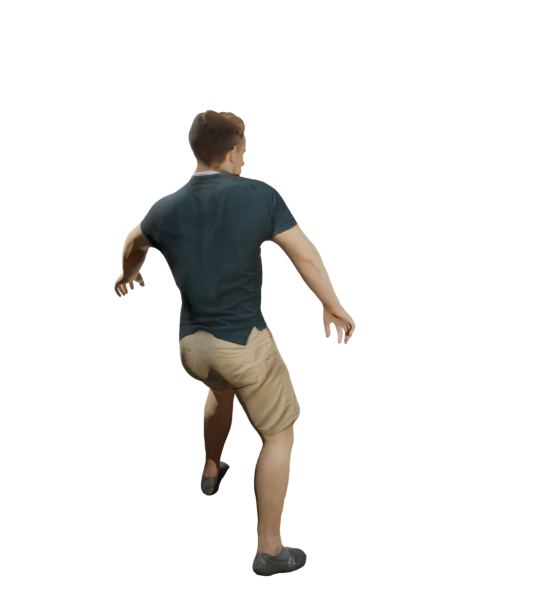}
            & \includegraphics[width=20mm]{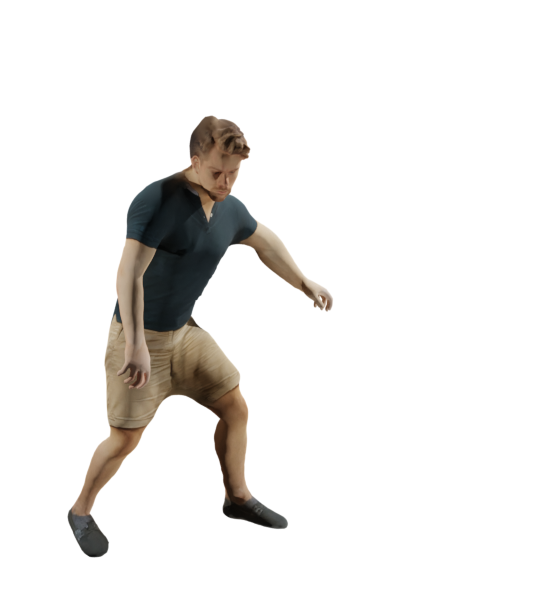}
            & \includegraphics[width=20mm]{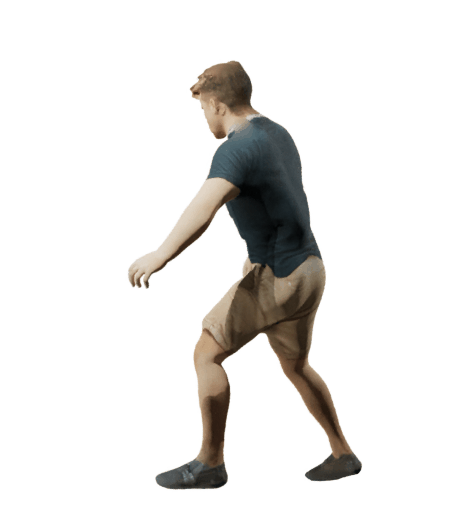}
            & \includegraphics[width=20mm]{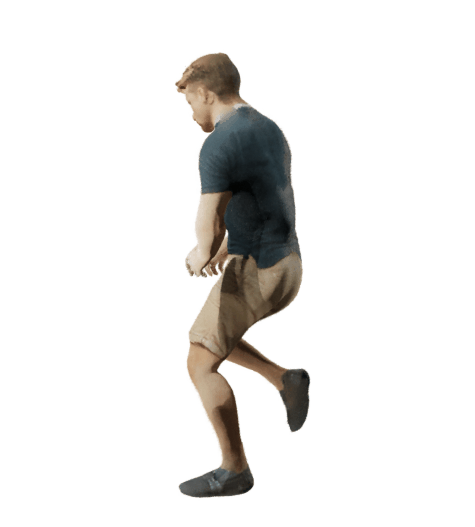}
            & \includegraphics[width=20mm]{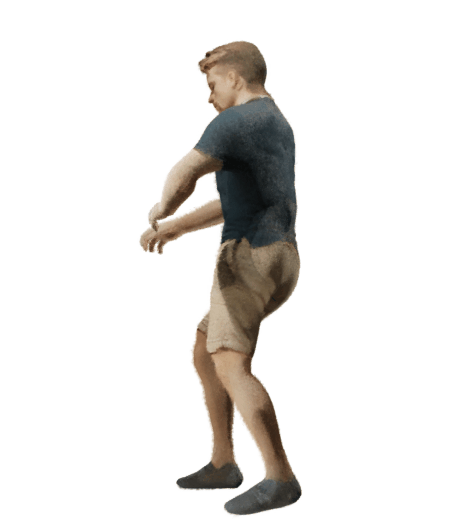}
            & \includegraphics[width=20mm]{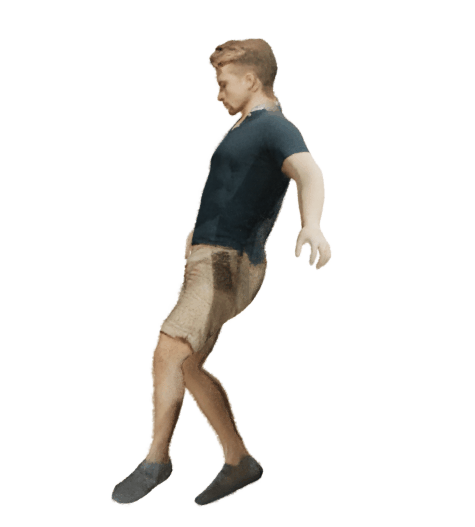}
            & \includegraphics[width=20mm]{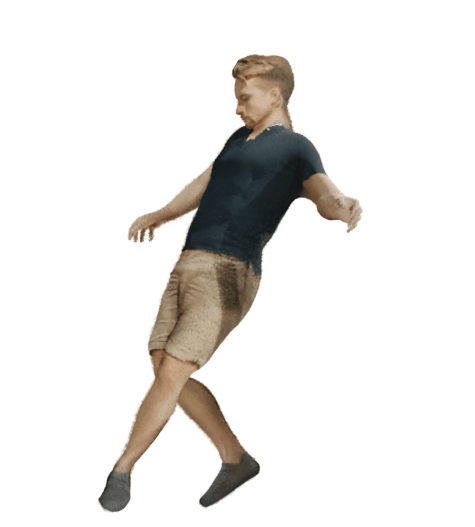}
            & \includegraphics[width=20mm]{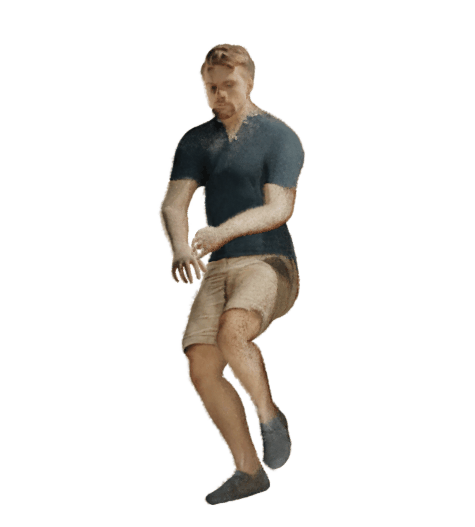}
            & \includegraphics[width=20mm]{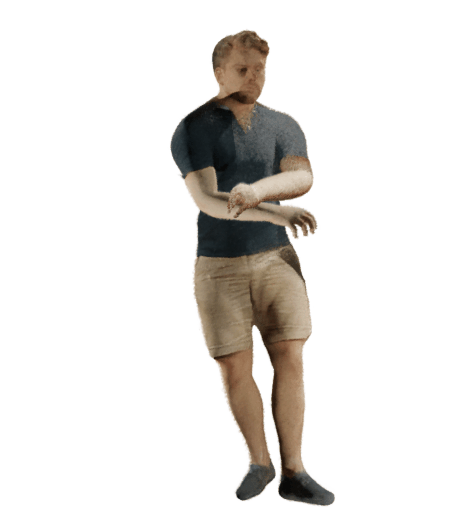}
            & \includegraphics[width=20mm]{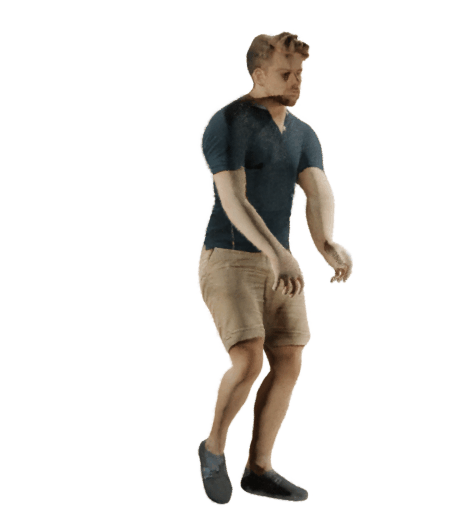}\\
             \includegraphics[width=20mm]{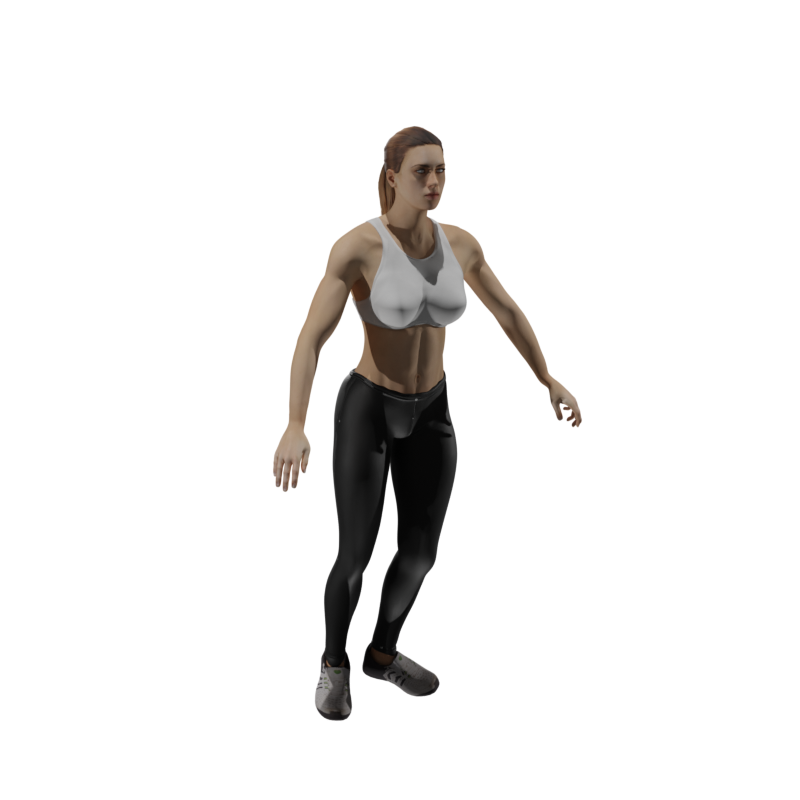}
            & \includegraphics[width=20mm]{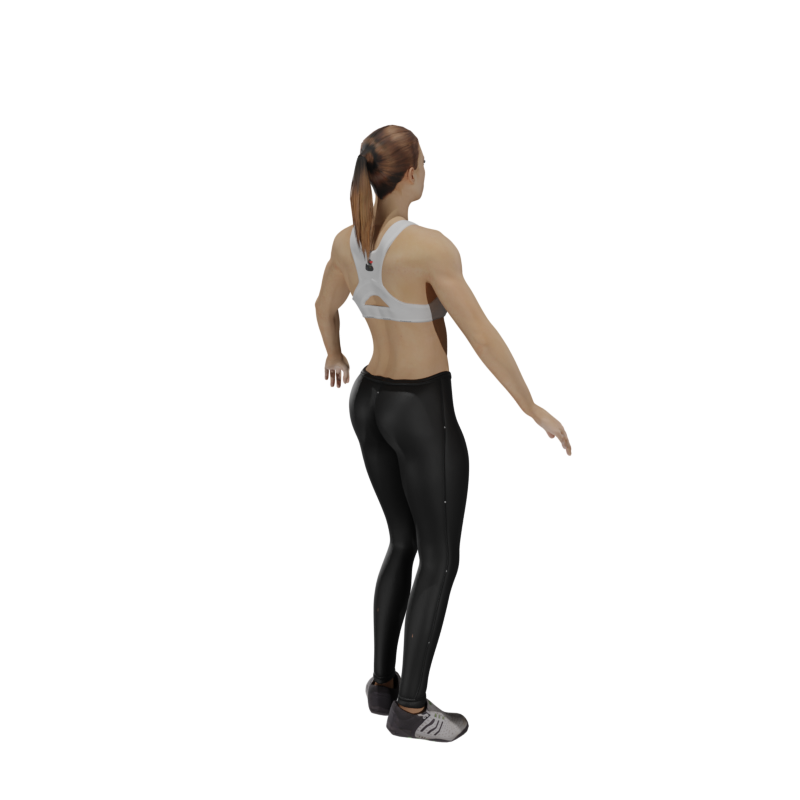}
             & \includegraphics[width=20mm]{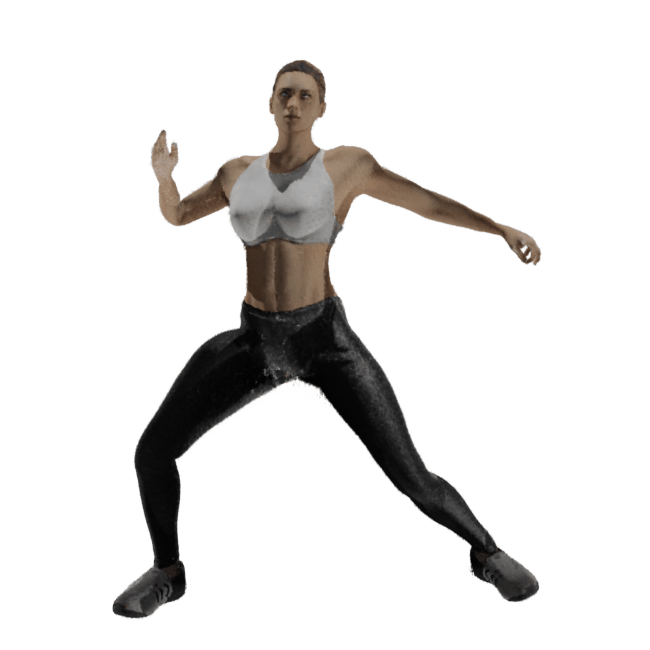}
             & \includegraphics[width=20mm]{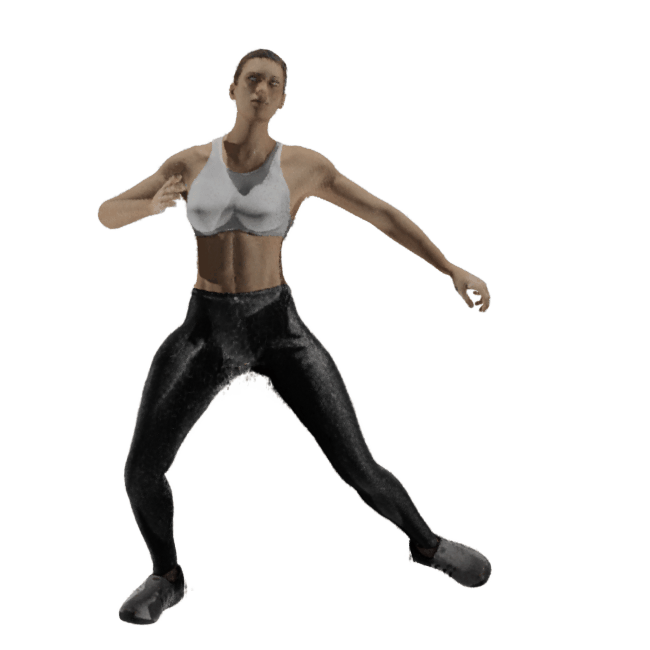}
            & \includegraphics[width=20mm]{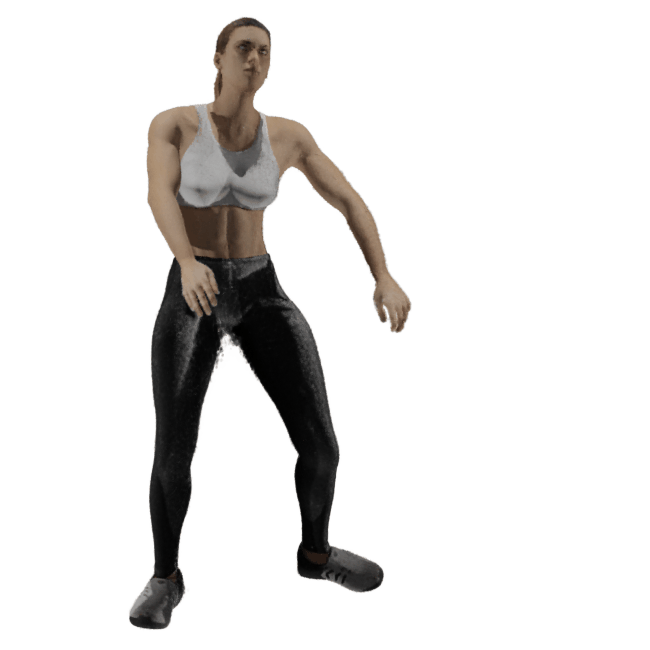}
            & \includegraphics[width=20mm]{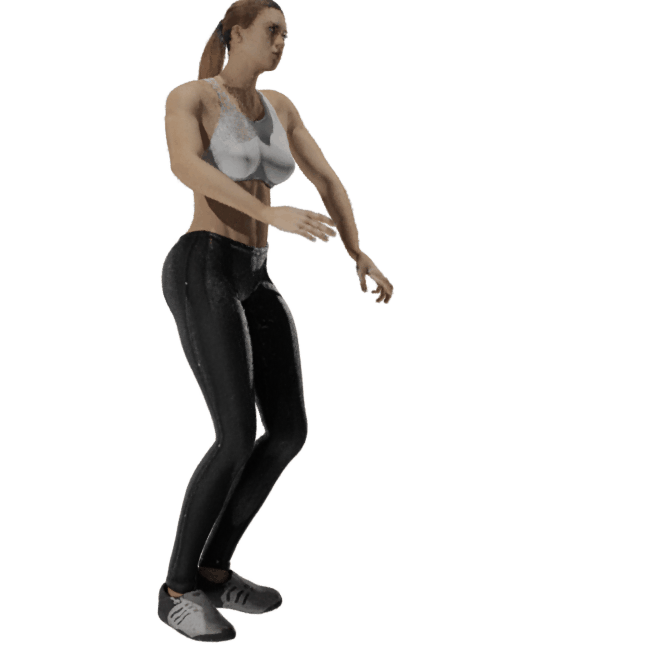}
             & \includegraphics[width=20mm]{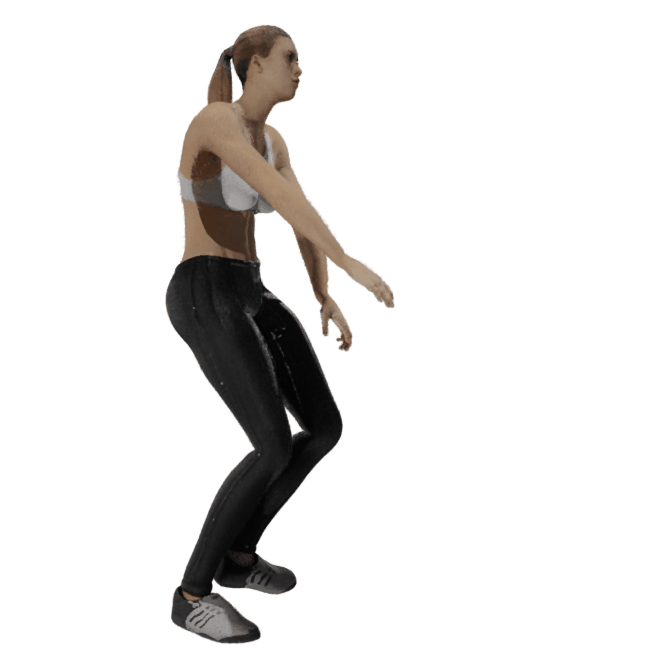}
             & \includegraphics[width=20mm]{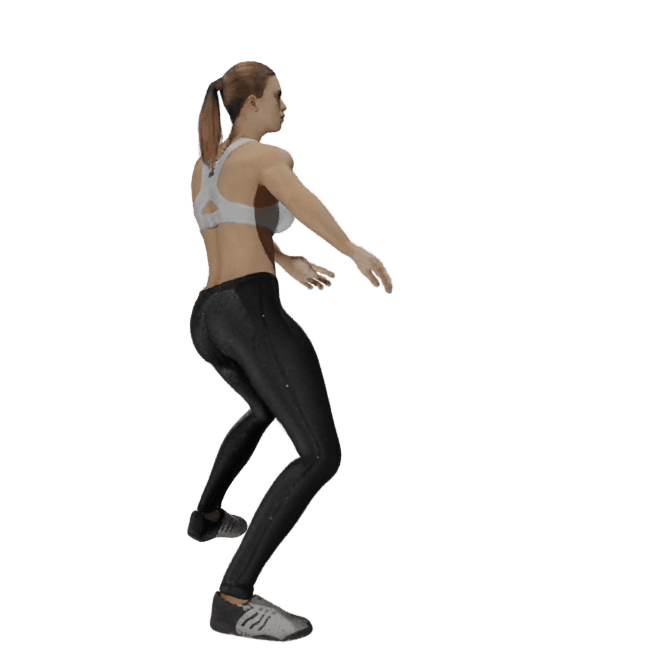}
            & \includegraphics[width=20mm]{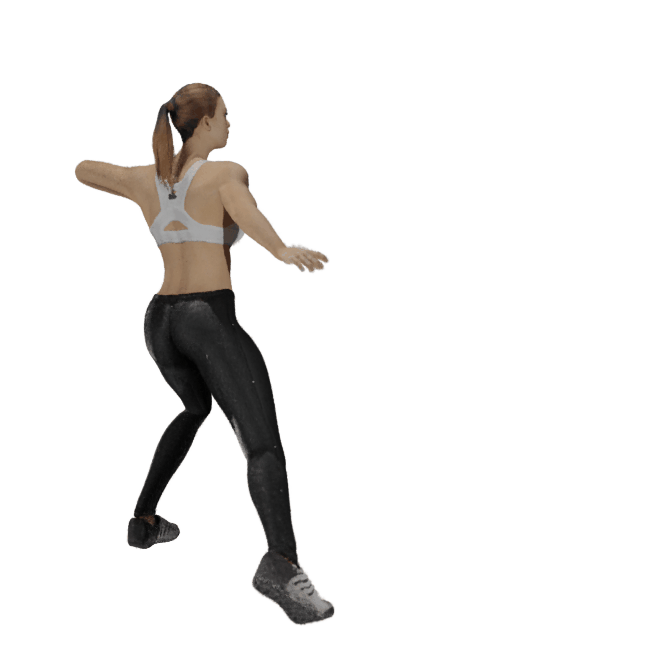}
            & \includegraphics[width=20mm]{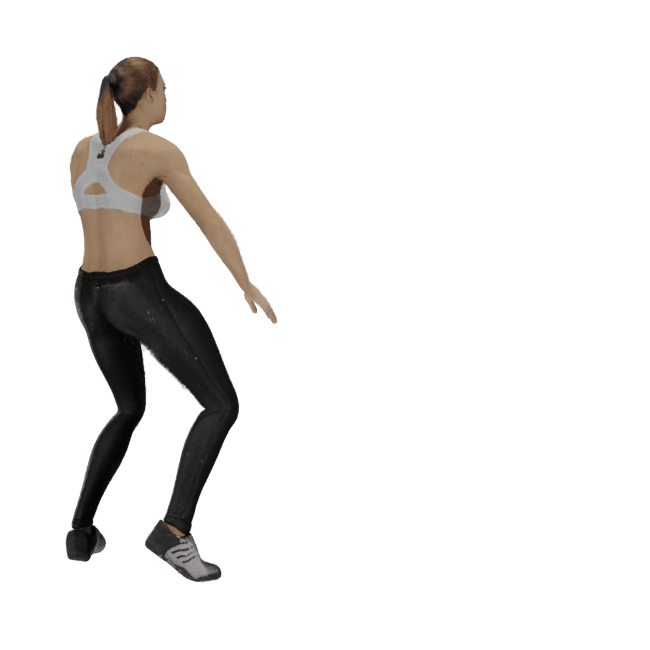}\\
              \includegraphics[width=20mm]{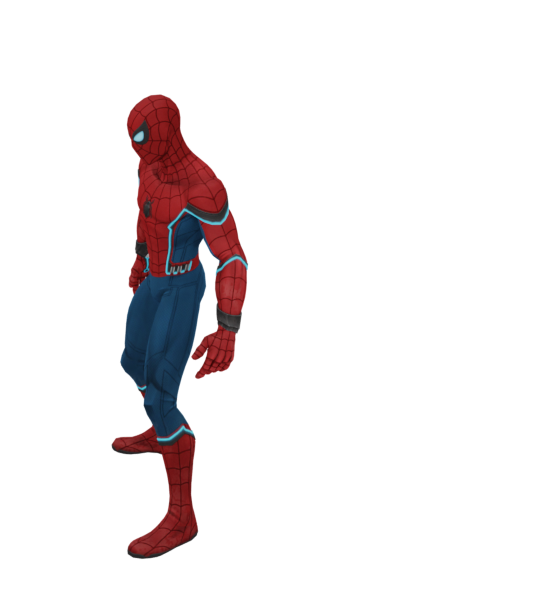}
            & \includegraphics[width=20mm]{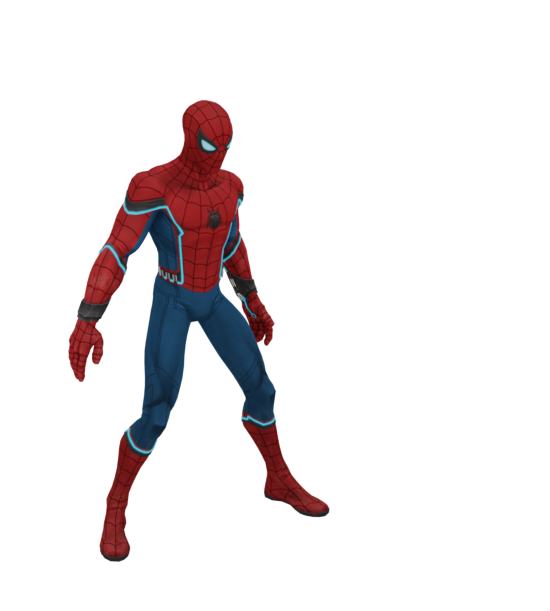}
            & \includegraphics[width=20mm]{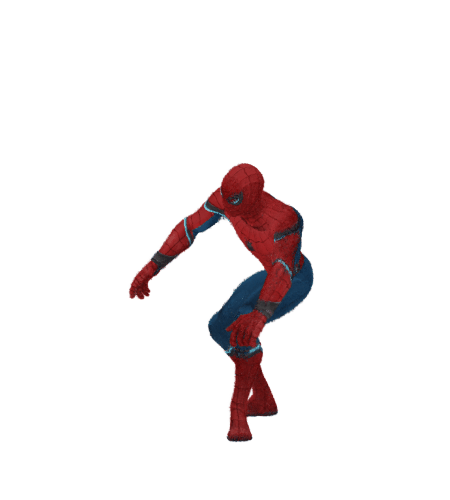}
             & \includegraphics[width=20mm]{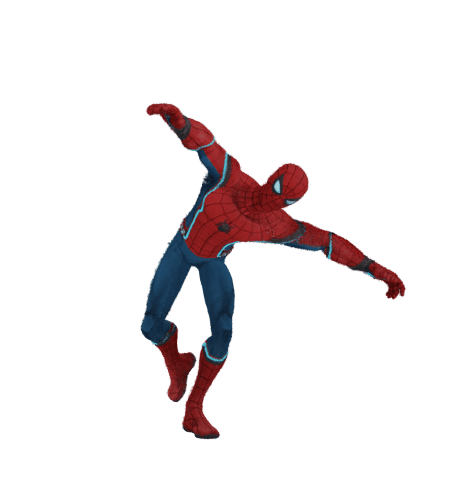}
            & \includegraphics[width=20mm]{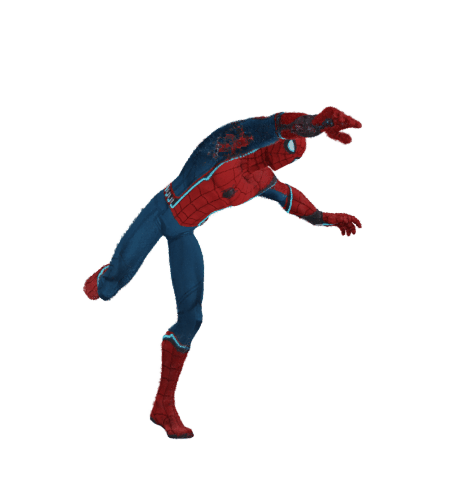}
             & \includegraphics[width=20mm]{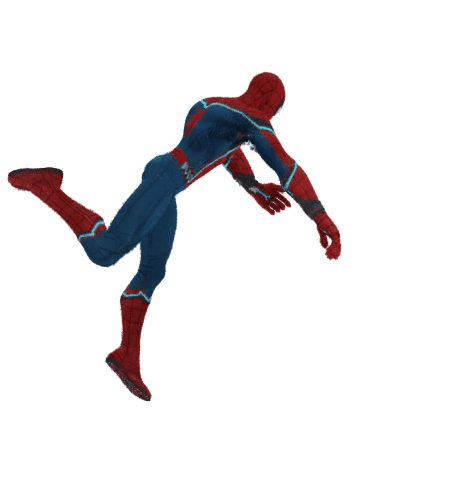}
             & \includegraphics[width=20mm]{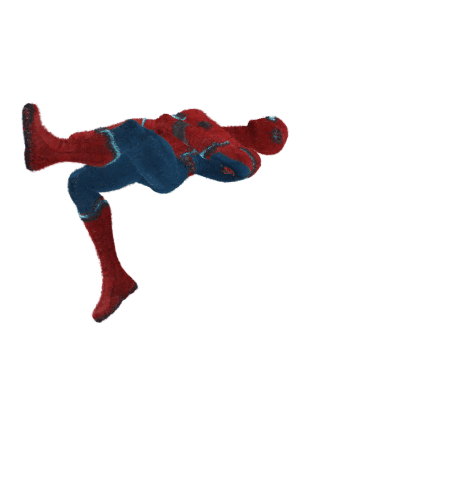}
            & \includegraphics[width=20mm]{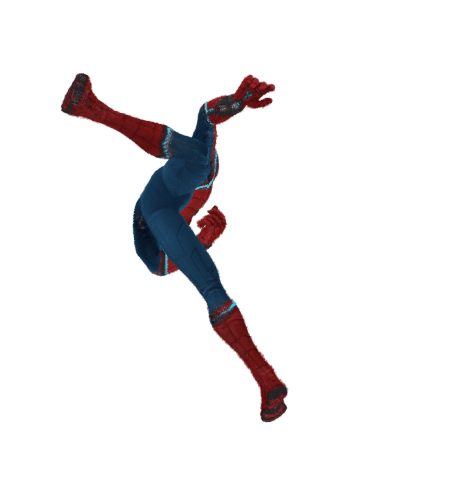}
            & \includegraphics[width=20mm]{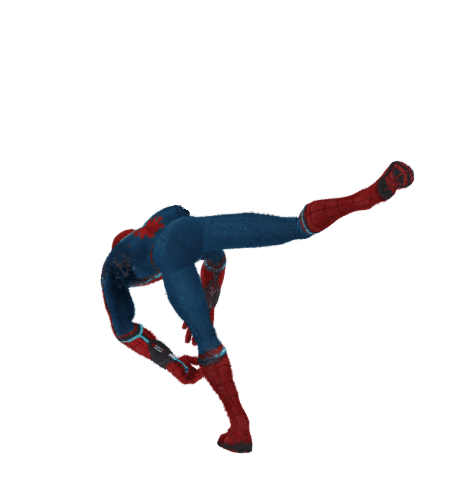}
            & \includegraphics[width=20mm]{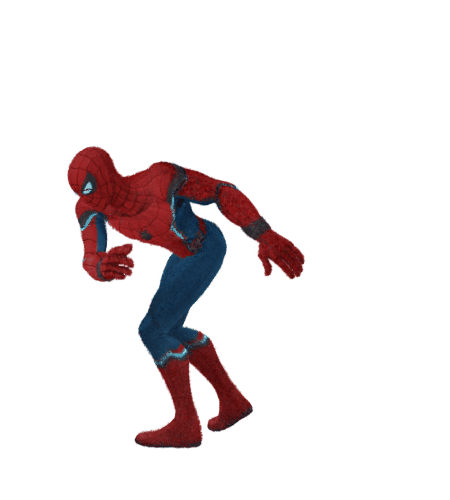}\\
        \end{tabular}%
    \caption{Experiment results of human motion synthesis. The left two images of each column represent the multi-view images of the static human, and the right columns are the novel views of unseen motions.  }
    \label{fig:human_result}
\end{figure*}

\begin{figure*}[h]
    \centering
        \begin{tabular}{p{16mm}p{20mm}|p{16mm}p{16mm}p{16mm}p{16mm}p{16mm}p{16mm}p{16mm}}
             \includegraphics[width=25mm]{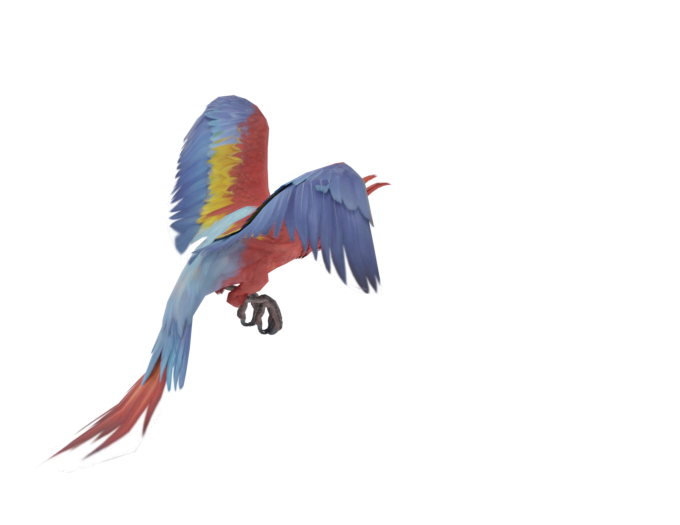}
            & \includegraphics[width=25mm]{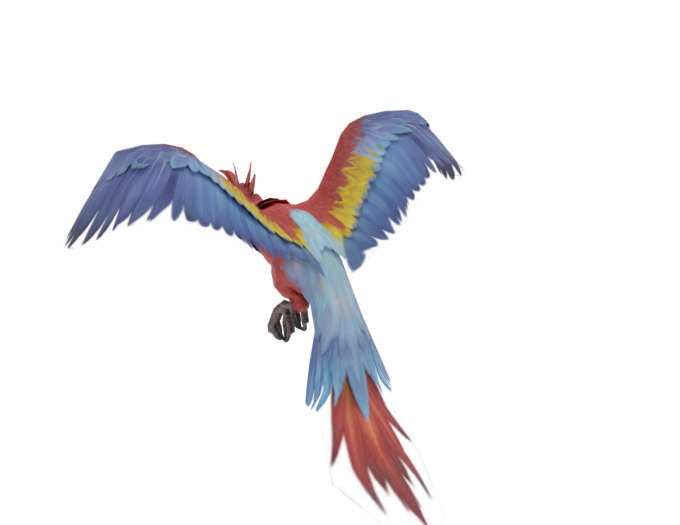}
            & \includegraphics[width=25mm]{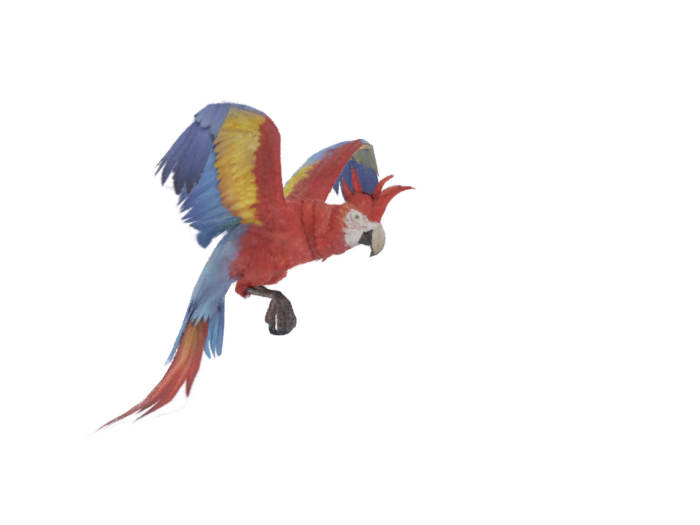}
            & \includegraphics[width=25mm]{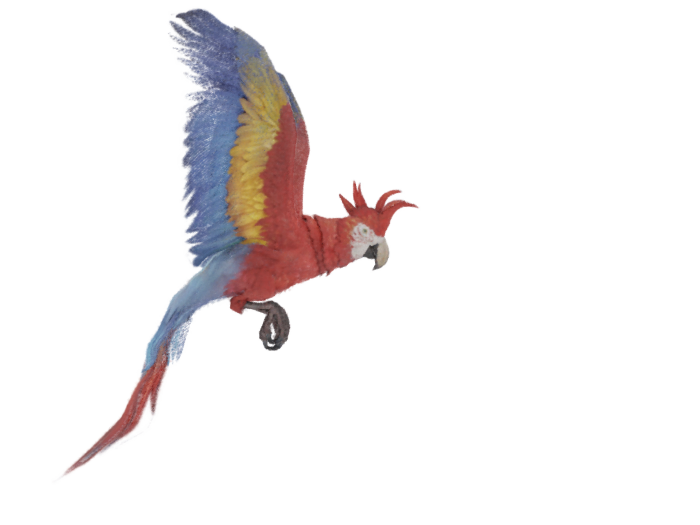}
            & \includegraphics[width=25mm]{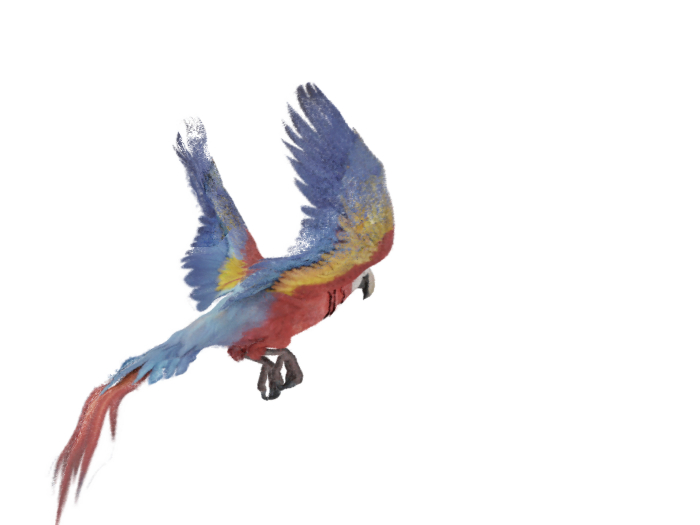}
            & \includegraphics[width=25mm]{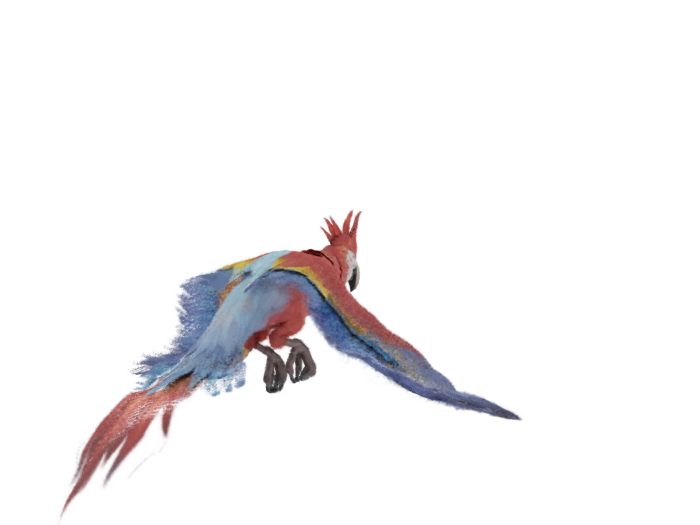}
            & \includegraphics[width=25mm]{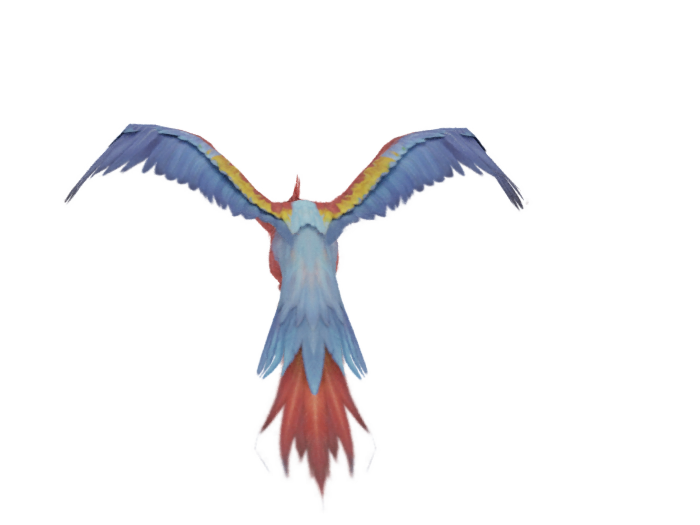}
            & \includegraphics[width=25mm]{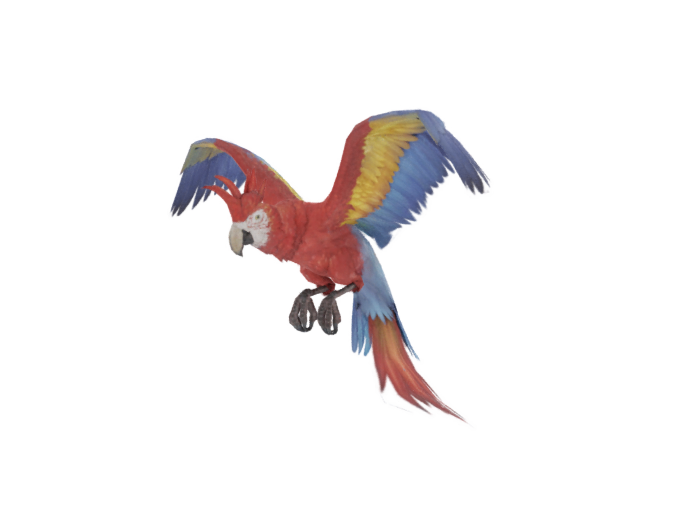}\\
            \includegraphics[width=25mm]{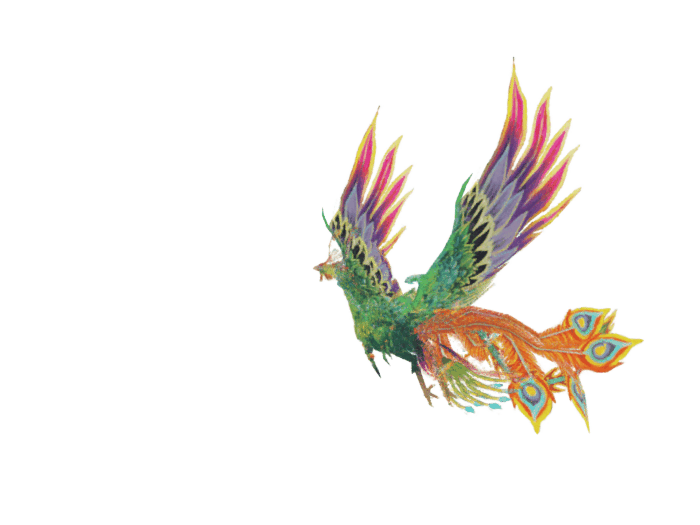}
            & \includegraphics[width=25mm]{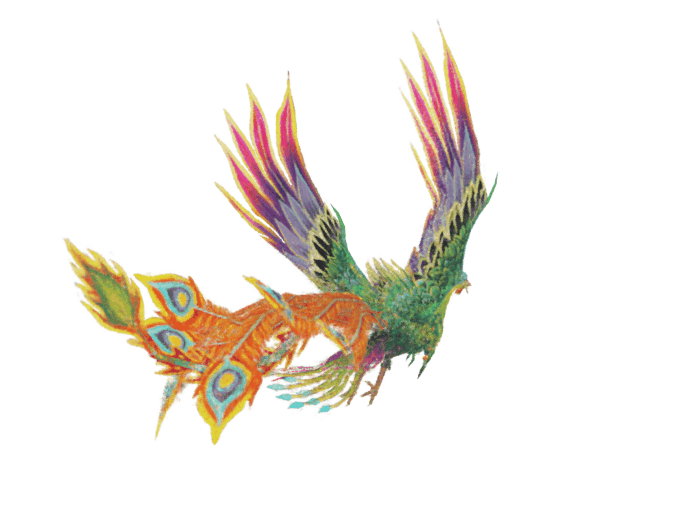}
            & \includegraphics[width=25mm]{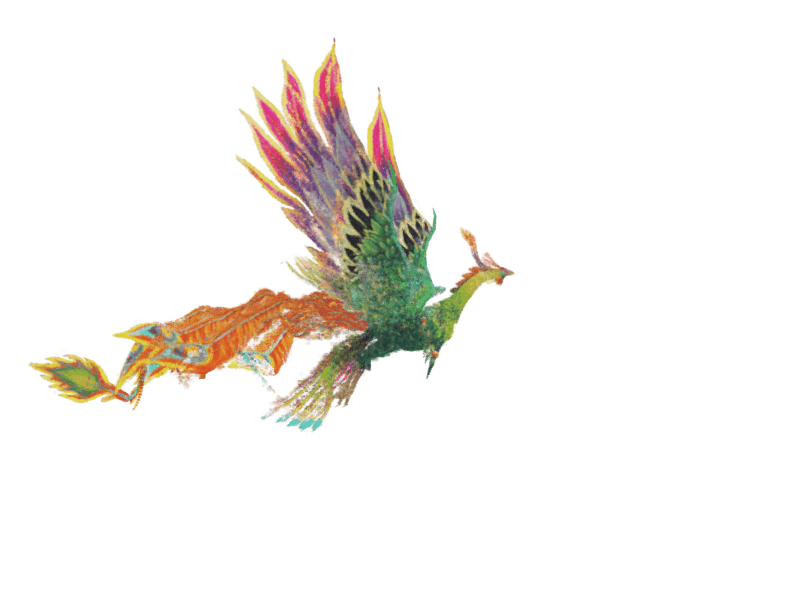}
            & \includegraphics[width=25mm]{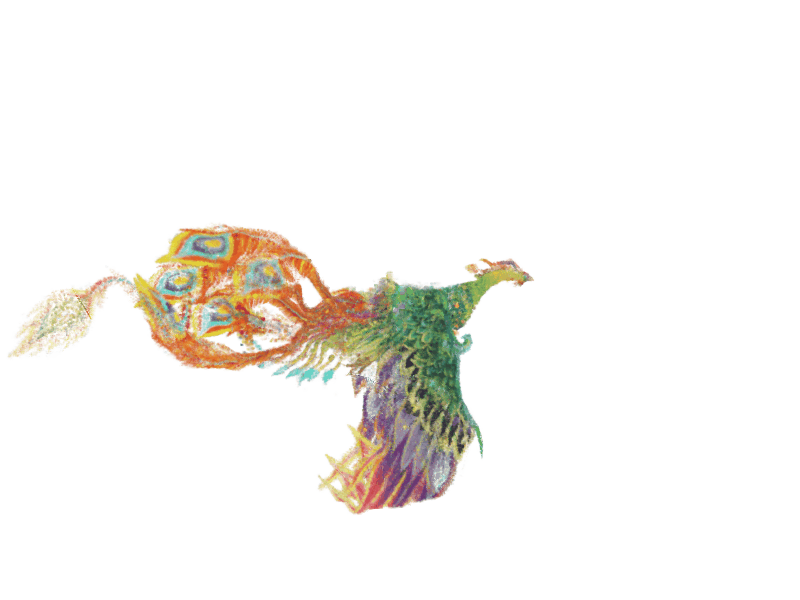}
            & \includegraphics[width=25mm]{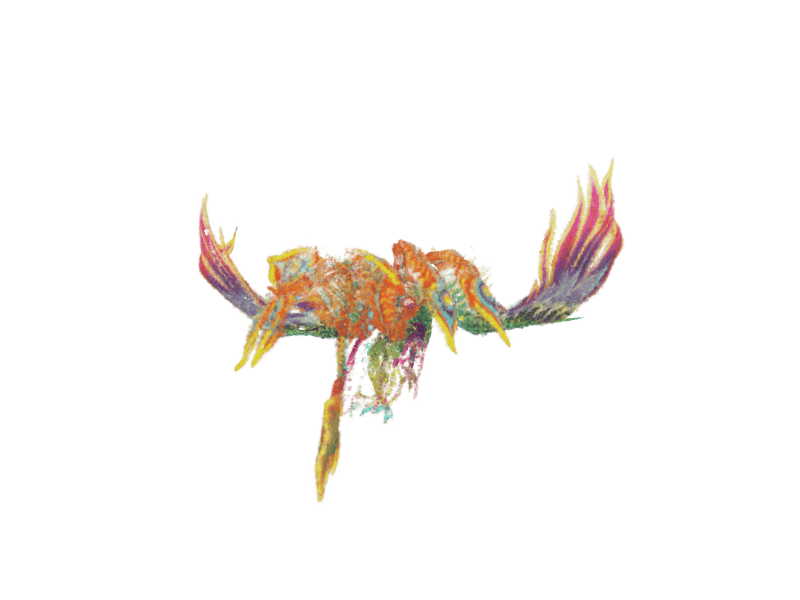}
            & \includegraphics[width=25mm]{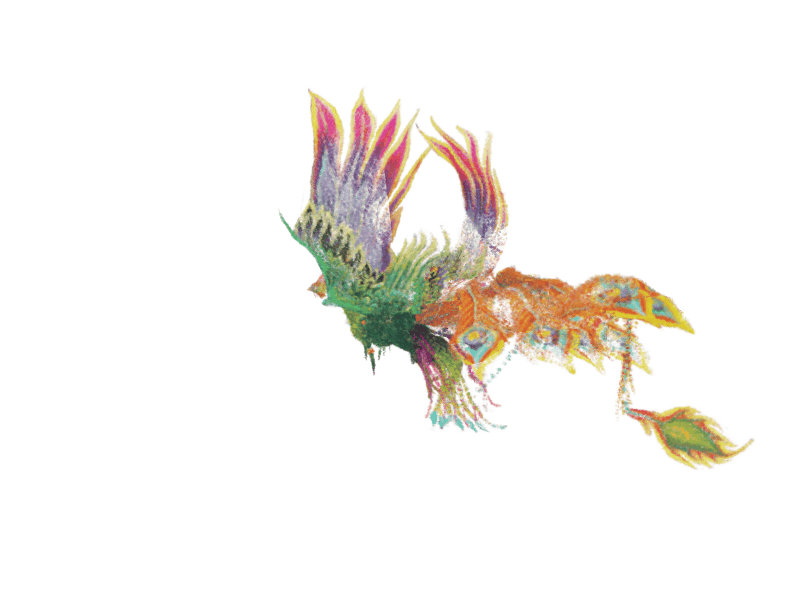}
            & \includegraphics[width=25mm]{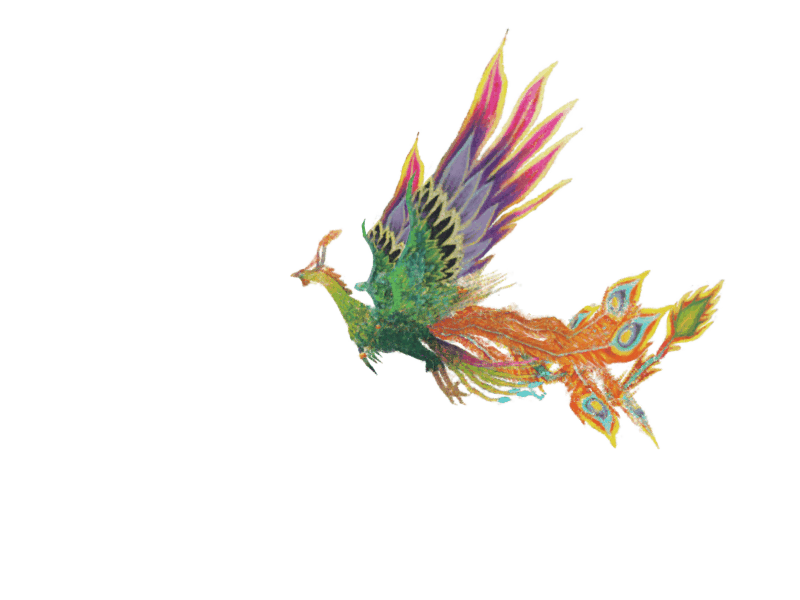}
            & \includegraphics[width=25mm]{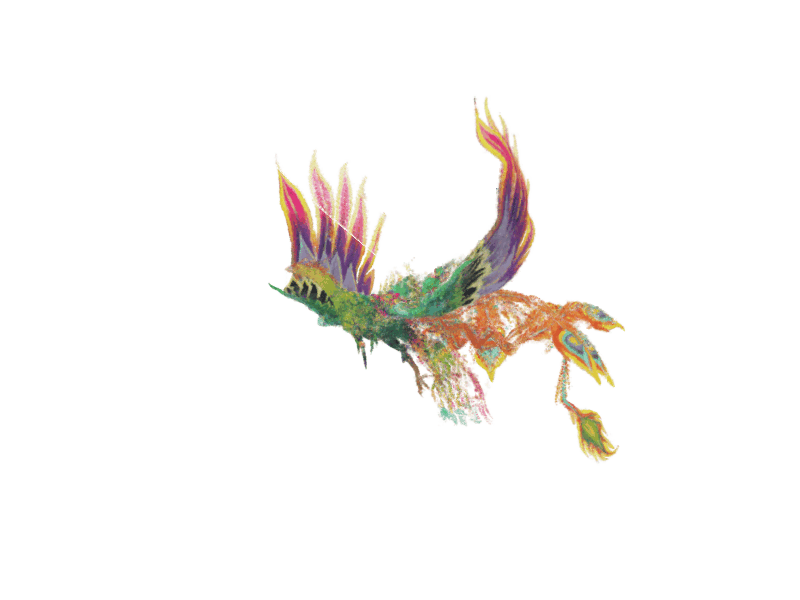}\\
        \end{tabular}%
    \caption{Experiment Results on character motion synthesis. The left two images of each column represent the multi-view images of the static character, and the right columns are the novel views of unseen motions.}
    \label{fig:character_result}
\end{figure*}

\section{Experiments}
We present the performance of our method on several synthesis datasets. 
In the remainder of this section, we first introduce our
experiment settings. Then we visualize the results of our experiments. Finally, we show some comparison and failure cases and analyze the performance.
\subsection{Experiment Settings}

\textbf{Dataset.} 
Due to the lack of publicly available benchmarks for human motion synthesis datasets, we create synthetic humanoid datasets constructed by BlenderNeRF\cite{Raafat_BlenderNeRF_2023}. Specifically, we download 3D animated models and use BlendNeRF to generate camera views. 
We first take one static pose of the human as canonical space and play the camera with the sphere method operator to export 100 camera views and their corresponding rendered images. We use these multi-view static scenes as \textit{training dataset} to train PointNeRF. 
Then we make the humanoid movable and export the mesh vertices of each frame as the animated key points for DPF \cite{prokudin2023dynamic}. 
Afterward, we play the camera on a horizontal circle while the humanoid is moving (1 view per frame) as \textit{test dataset} and export the camera views and rendered images as ground truth.
Apart from human datasets, we also generate character datasets like Robot, Turtle, Dragons, etc. to test the generalization ability of our method. Note that in our settings, all of the above datasets are without background and exposed to ambient environment light.
We also download static indoor models and add them as the background to our human datasets. We followed the same pipeline to generate datasets with background, in order to demonstrate the ability of our methods to be applied to these scenes as well.

\textbf{Implementation details.}
We first pre-train the PointNeRF using the multi-view static scene training data. We train for 200000 iterations on NVIDIA 3090 and get the point cloud as canonical space $X_{src}$ for deformation. In the second stage, we take the point cloud $X_{src}$ and key points $V_{src}$ in canonical space and key points of new poses $X_{trg}$ to train DPF for 2000 iterations each frame to get the deformed point cloud $X_{trg}$. Then we use the PointNeRF Rendering module combined with our ray-bending method to get the rendered images for the deformed point cloud $X_{trg}$. 

\textbf{Evaluation metric.} 
We measure the reconstruction and deformation quality with Peak Signal-to-Noise Ratio (PSNR). We project the 3D bounding box of the human body onto the image plane to get a 2D mask and only calculate the PSNR of the mask area instead of the whole image. 
For the animation synthesis results, we provide qualitative visual results in Fig. \ref{fig:human_result}, Fig. \ref{fig:character_result} and \href{https://youtu.be/3iZ_89IwZUU}{videos}.

\begin{table}[h]
\centering
\begin{subtable}[ht]{\columnwidth}
\resizebox{\columnwidth}{!}{
\begin{tabular}[t]{lccccc}
\hline
& Man & Woman & Samba & Gangnam & Spiderman\\
\hline
Static & 38.778 & 30.837 & 35.474 & 37.662 & 35.711 \\
Deformed & 10.061 & 19.163 & 24.265 & 14.112 & 20.373 \\
\hline
\end{tabular}
}
\caption{PSNR on Human Dataset.\label{tab:comp_static_deform_human}}
\end{subtable}
\begin{subtable}[ht]{\columnwidth}
\resizebox{\columnwidth}{!}{
\begin{tabular}[t]{lccccc}
\hline
& Robot & Parrot & Turtle & Dragon & Phoenix\\
\hline
Static & 31.607 & 31.898 & 41.422 & 42.242 & 32.163\\
Deformed & 14.41 & 10.501 & 25.087 &11.450&11.550 \\
\hline
\end{tabular}
}
\caption{PSNR on character Dataset.\label{tab:comp_static_deform_character}}
\end{subtable}
\caption{PSNR Comparison for deformed results and static results (averaged on the whole animation sequence).}
\label{tab:comp_static_deform}
\end{table}%

\subsection{Experiment Results}
\textbf{Animation.}
Given a static point cloud in the canonical space trained by PointNeRF and a set of key points of a human performing unseen arbitrary motions, our method can synthesize novel pose images of the human performing these motions without training. As shown in Fig. \ref{fig:human_result}, the left two images of each column represent the multi-view images of the static human reconstructed by PointNeRF in canonical space, and the right  are the novel images of unseen motions. Our method demonstrates excellent visual quality on motions with large displacements and rotations. We also show the quantitative results in Table. \ref{tab:comp_static_deform_human} to compare the PSNR between the static humans and the deformed ones.

\begin{figure*}[ht]
    \centering
        \begin{tabular}{p{12mm}p{16mm}|p{10mm}p{11mm}p{12mm}p{10mm}p{11mm}p{12mm}p{10mm}p{11mm}p{12mm}}
             \multicolumn{2}{c}{$N_{kp}=20194/347834$} & \multicolumn{3}{c}{$N_{kp}=20$} & \multicolumn{3}{c}{$N_{kp}=200$} & \multicolumn{3}{c}{$N_{kp}=2000$}\\
             \includegraphics[width=20mm]{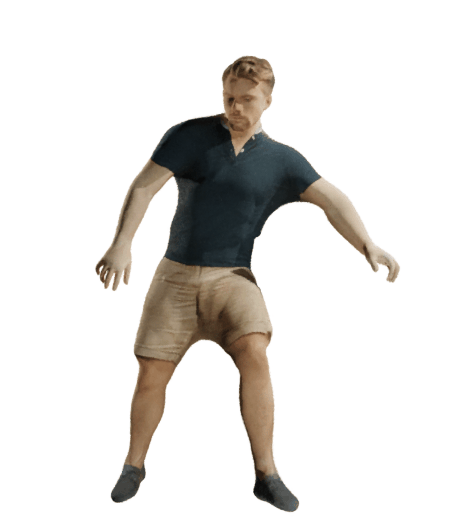}
            & \includegraphics[width=20mm]{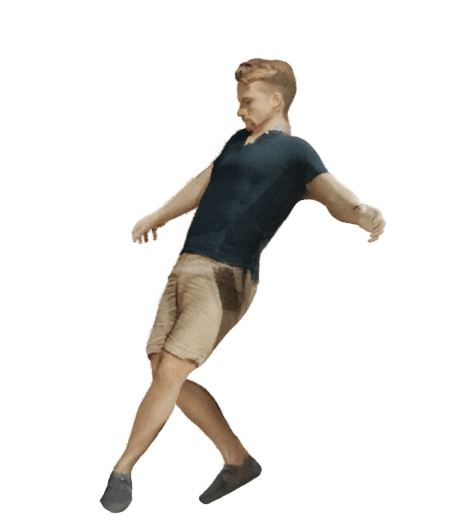}
            & \includegraphics[width=18mm]{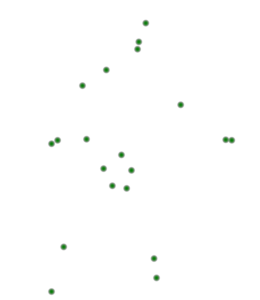}
            & \includegraphics[width=18mm]{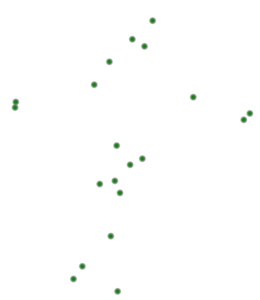}
            & \includegraphics[width=20mm]{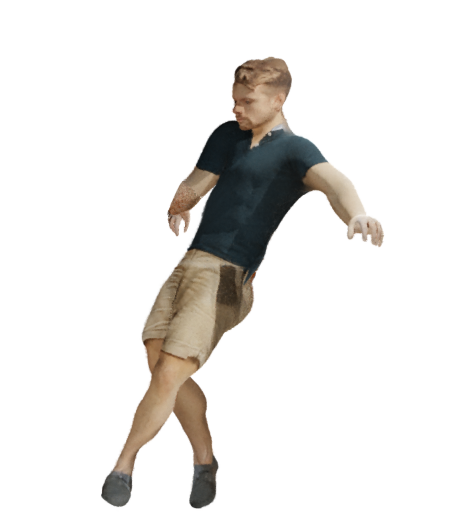}
            & \includegraphics[width=18mm]{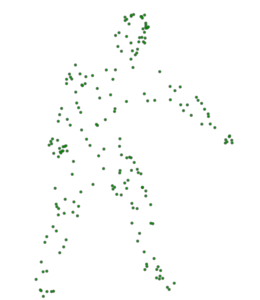}
            & \includegraphics[width=18mm]{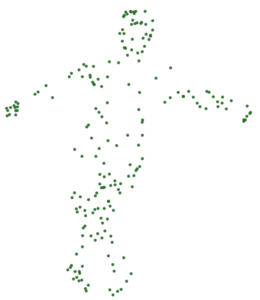}
            & \includegraphics[width=20mm]{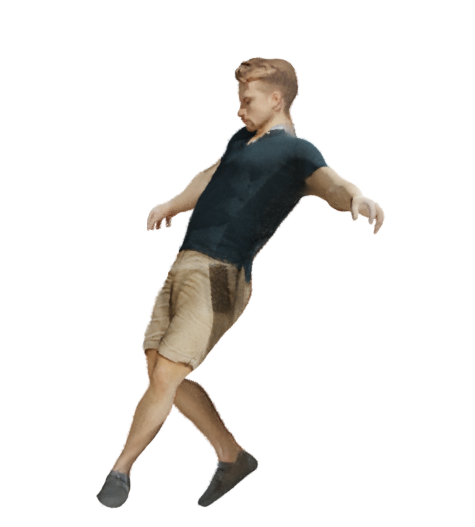}
            & \includegraphics[width=18mm]{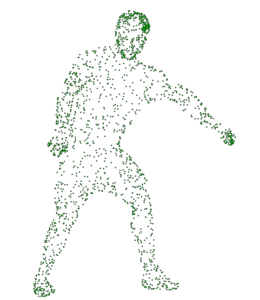}
            & \includegraphics[width=18mm]{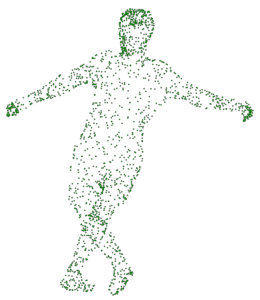}
            & \includegraphics[width=20mm]{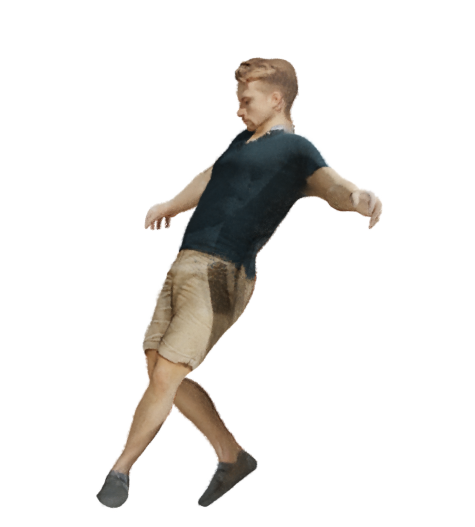}\\
             \multicolumn{2}{c}{$N_{kp}/N_{pc}=7466/315837$} & \multicolumn{3}{c}{$N_{kp}=20$} & \multicolumn{3}{c}{$N_{kp}=200$} & \multicolumn{3}{c}{$N_{kp}=2000$}\\
            \includegraphics[width=20mm]{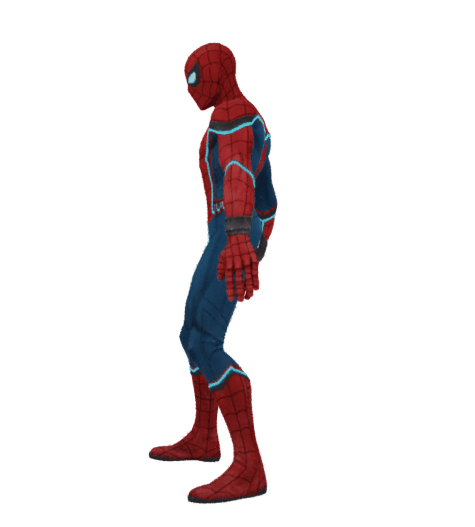}
            & \includegraphics[width=20mm]{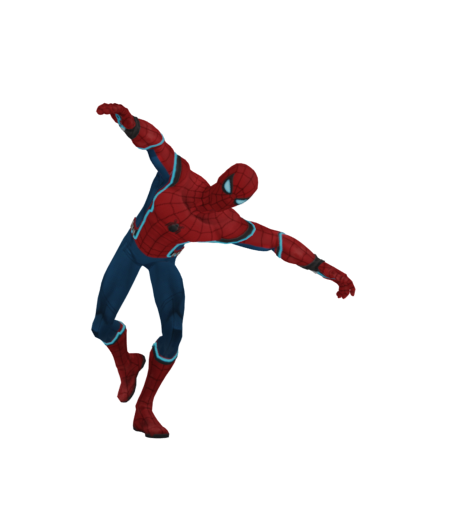}
            & \includegraphics[width=18mm]{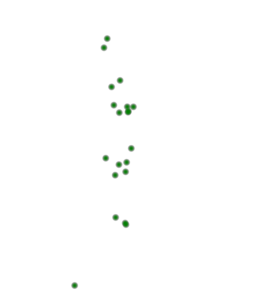}
            & \includegraphics[width=18mm]{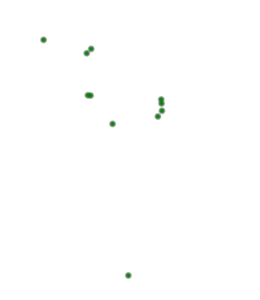}
            & \includegraphics[width=20mm]{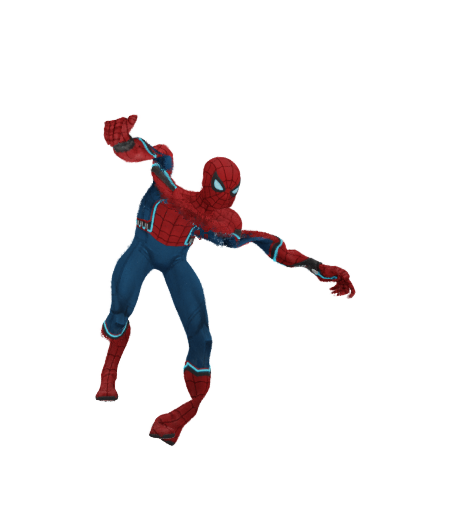}
            & \includegraphics[width=18mm]{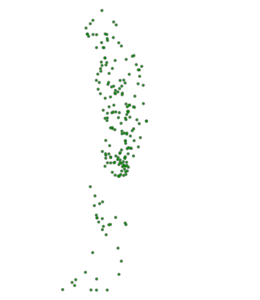}
            & \includegraphics[width=18mm]{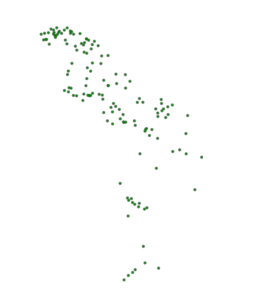}
            & \includegraphics[width=20mm]{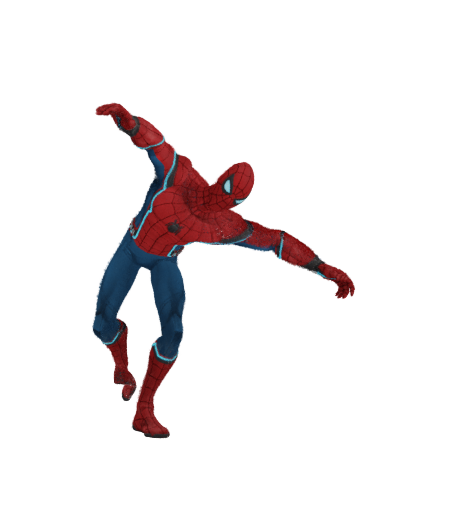}
            & \includegraphics[width=18mm]{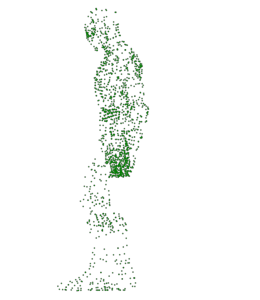}
            & \includegraphics[width=18mm]{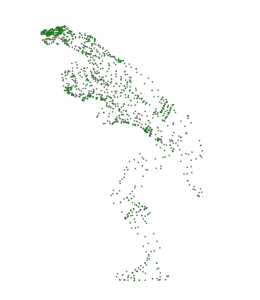}
            & \includegraphics[width=20mm]{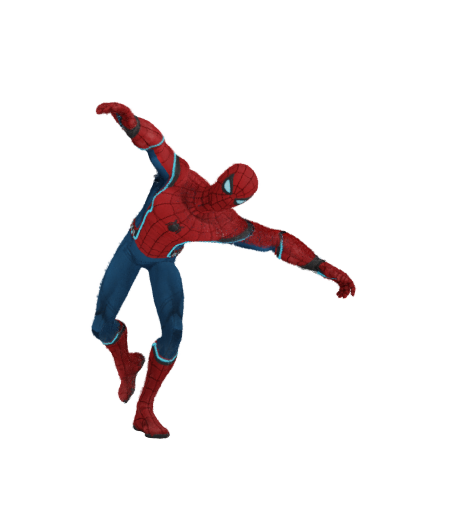}\\
        \end{tabular}%
    \caption{Keypoint number comparison. $N_{kp} / N_{pc}$ denotes the total mesh vertices vs. the total number of points of the point cloud reconstructed by PointNeRF, and $N_{kp}=i (i=20, 200, 2000)$ refers to the number of key points used for deformation. In each subfigure, the left point cloud is the key points of the canonical pose and the right point cloud is the key points of the deformed pose. Note that the poses of key points look different from the images because they are projected to different camera views, but still they actually represent the same poses.}
    \label{fig:keypoint_result}
\end{figure*}

\textbf{Generalization.}
In the generalization test, we evaluate our approach on different datasets across different subjects. Since our method does not rely on any prior model, such as SMPL \cite{SMPL:2015}, our model is assumed to be not limited to humanoids. we extended the task of human synthesis to character synthesis to validate our method's generalization ability. We conducted experiments on the Turtle, Robot, Dragon, Parrot, and Phoenix datasets using the same method and show visual results in Fig. \ref{fig:character_result} and quantitative results in Table. \ref{tab:comp_static_deform_character}. Even though these datasets are significantly different in appearance, our method still can achieve impressive results.

\subsection{Evaluation}
\textbf{Ablation Study.} We conduct the ablation study to compare the ray-bending results with no-bending results. Additionally, we also explore the impact of keypoint numbers for deformation.

\textit{Ray-bending.}
We compared the visual quality and quantitative improvement of our method with and without ray-bending. Table. \ref{tab:psnr_comp_bending} shows the comparison of PSNR values, indicating that our method with and without ray-bending shows similar results. Ray-bending did not show significant differences in PSNR value because most of our objects are exposed to ambient environment light and have surfaces made of diffuse materials, where no detailed view-dependent information be learned. Thus, the effectiveness of ray-bending cannot be shown on this coarse level construction.

\begin{figure}[h]
\begin{subfigure}[b]{\columnwidth}
    \includegraphics[width=\columnwidth]{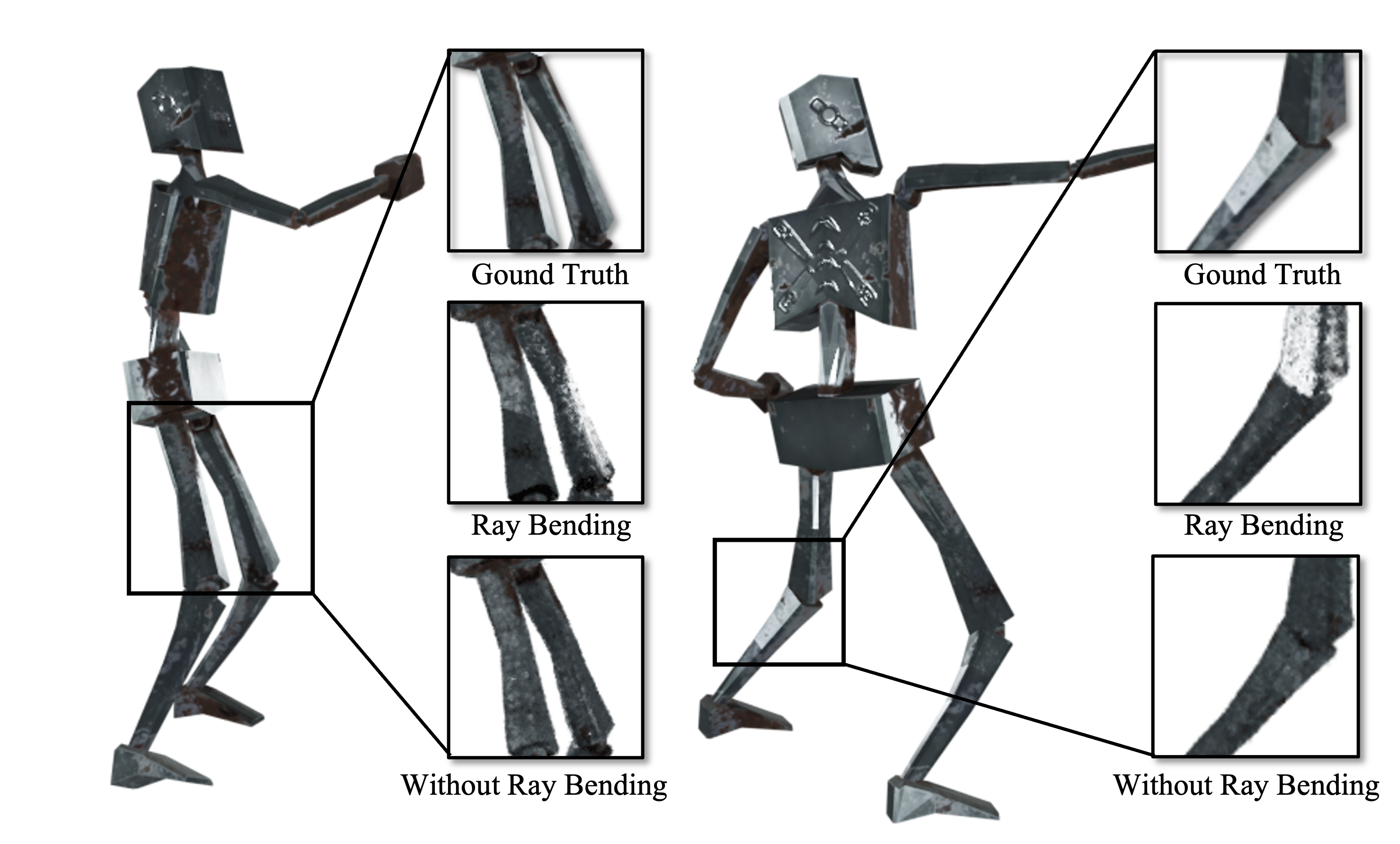}
    \caption{Raybending preserves specular reflection.\label{fig:detail_raybending_robot}}
\end{subfigure}
\begin{subfigure}[b]{\columnwidth}
    \includegraphics[width=\columnwidth]{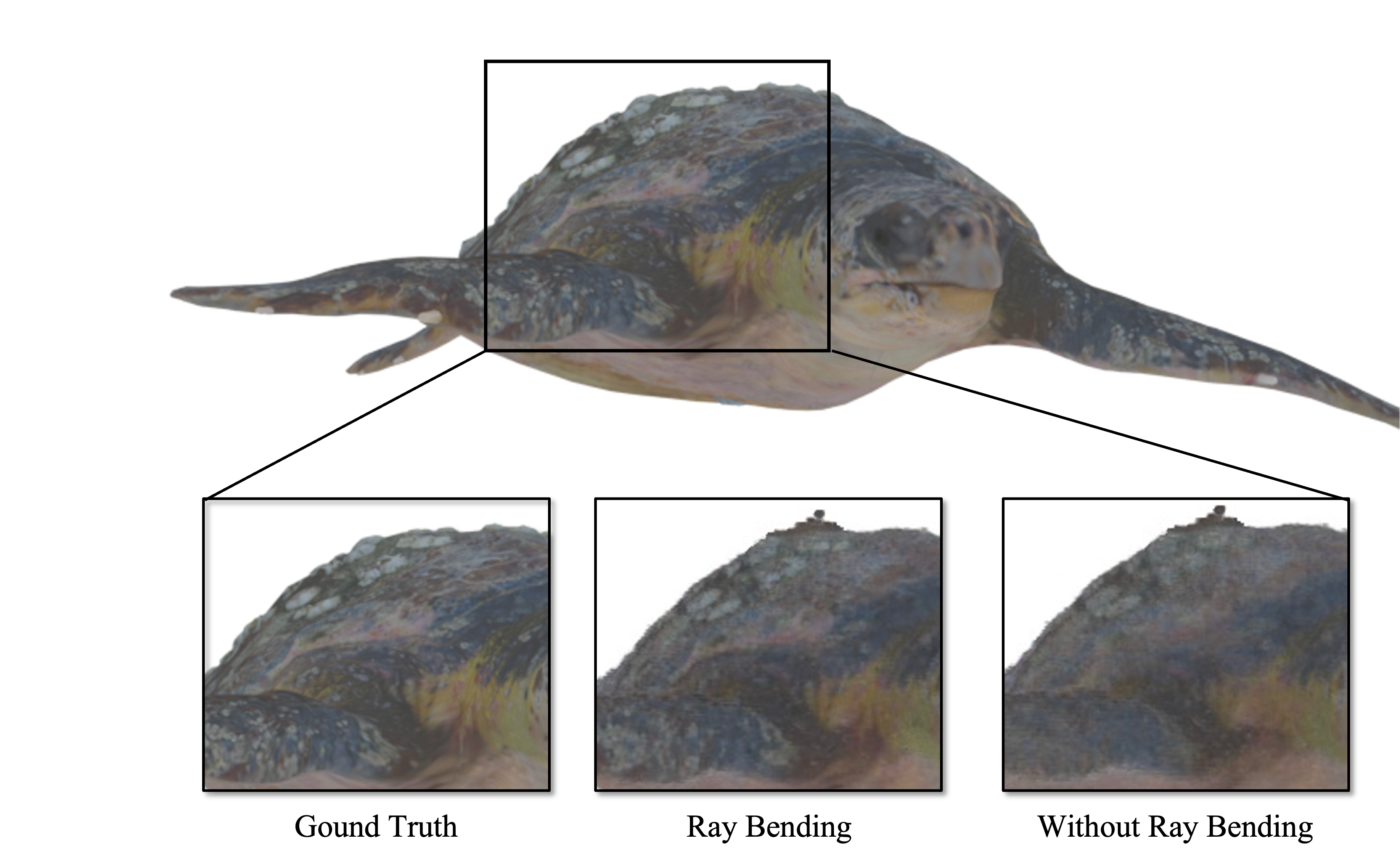}
    \caption{RayBending preserves texture.\label{fig:detail_raybending_turtle}}
\end{subfigure}
\caption{Visual details of the ablation studies on Ray-bending. The large images show the ground truth character and the highlighted regions compare the zoom-in details of ground truth, ray-bending, and no-bending results. }
\label{fig:detail_raybending}
\end{figure}

To demonstrate the validity of ray-bending, we design two cases that manage to manifest the conditional instability of finer-level rendering with no-bending approach. Firstly, we analyze the visual quality of the Robot dataset since the robot is made of specular material which is more easily affected by view changes. As shown in Fig. \ref{fig:detail_raybending_robot}, it can be observed that in the highlighted regions, our ray-bending method better reproduces the reflections, while the no-bending method fails to do so. The difference in position is considered to be caused by the light source position, which cannot be efficiently learned in PointNeRF. Secondly, we believe that no-bending rendering will lead to the direct query of weakly supervised view-dependent information under large deformation. In Fig. \ref{fig:detail_raybending_turtle}, we compare the improvement of ray-bending by flipping a turtle by 180 degrees. It can be seen that ray-bending better preserves the detailed texture of the turtle.

\begin{table}[ht]
\centering
\begin{subtable}[ht]{\columnwidth}
\resizebox{\columnwidth}{!}{
\begin{tabular}[t]{lccccc}
\hline
& Man & Woman & Samba & Gangnam & Spiderman\\
\hline
Ray-Bending & 10.061 & 19.163& 24.265 & 14.112 & 20.373 \\
No Bending & 9.794 &19.157 & 25.096  & 14.284 &  19.997\\
\hline
\end{tabular}
}
\caption{PSNR on Human Dataset.}
\end{subtable}
\begin{subtable}[ht]{\columnwidth}
\resizebox{\columnwidth}{!}{
\begin{tabular}[t]{lccccc}
\hline
& Robot & Parrot & Turtle & Dragon & Phoenix\\
\hline
Ray-Bending & 14.797 & 10.501 & 25.087 & 11.450& 11.549 \\
No Bending & 15.010 & 10.462 & 25.005 & 11.667&11.532 \\
\hline
\end{tabular}
}
\caption{PSNR on character Dataset.}
\end{subtable}
\caption{PSNR Comparison for ray-bending and no-bending.}
\label{tab:psnr_comp_bending}
\end{table}%

\textit{Number of Keypoints.}
To investigate the influence of the number of key points on deformation and explore the minimum number of key points that can be used, we varied the number of key points by randomly selecting 20, 200, and 2000 key points and compared the visual quality of the resulting deformations. Fig. \ref{fig:keypoint_result} illustrates the comparison of visual quality using different numbers of key points. As shown in the $N_{kp}=20$ case, too few key points will cause large distortions. When $N_{kp}=200$, the performance is largely improved. In our test, taking $N_{kp} \geq 300$ is enough to generate good deformation performance.

\begin{figure}[h]
\includegraphics[width=\columnwidth]{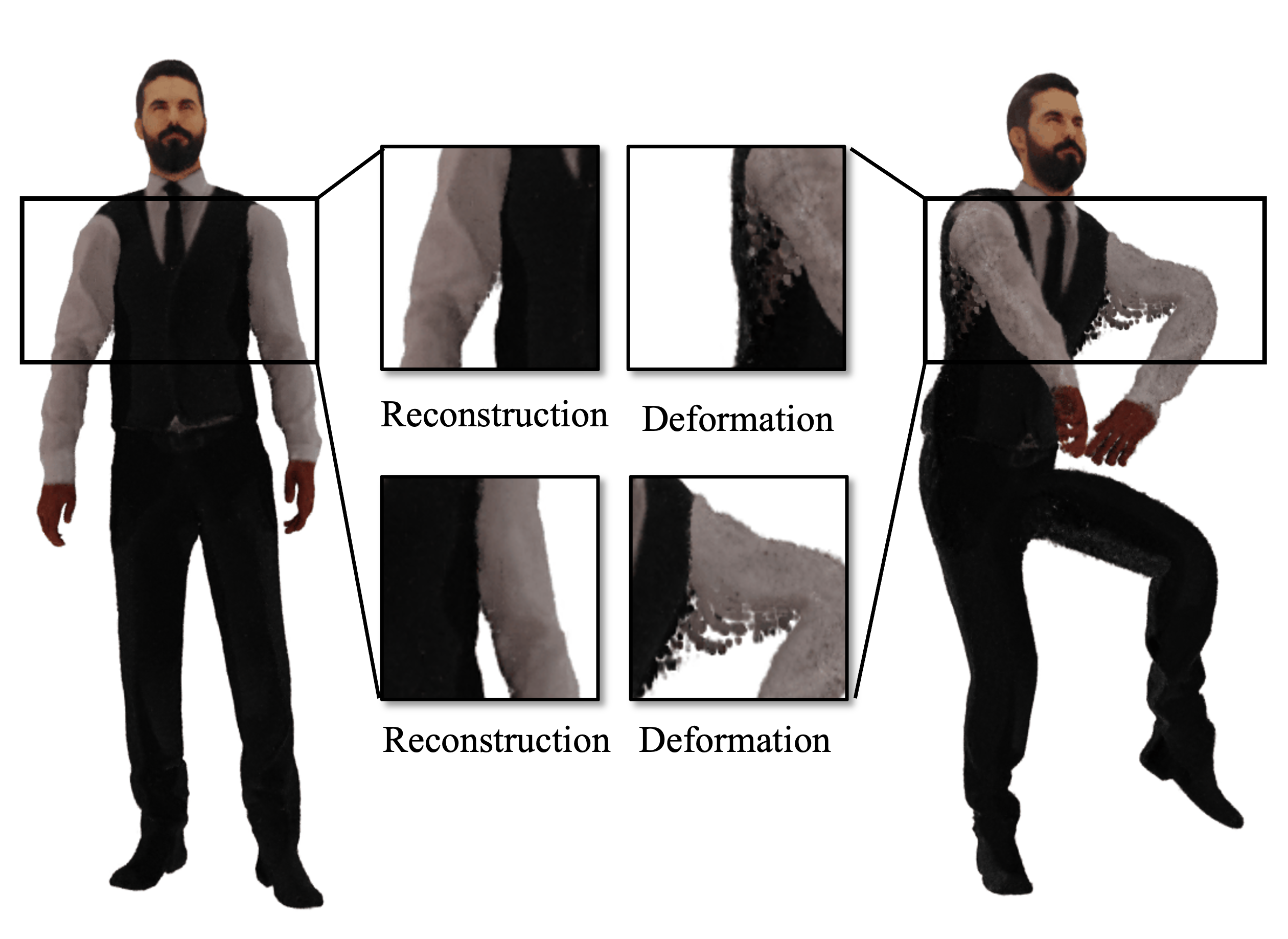}
\caption{An example of failure case. The zoom-in regions show the artefact of deformation.}
\label{fig:failure}
\end{figure}

\textbf{Failure Cases.}
Poor quality of the canonical space is one of the factors that led to the degraded synthesis results. When the canonical space contains noticeable shadows, occlusions, or regions not visible from the camera's perspective, PointNeRF will not reconstruct the local space and cause discontinuities or empty holes which limits the performance of DPF. For example, in Figure \ref{fig:failure}, the highlighted area of Gangnam in the canonical space was not reconstructed properly due to occlusion. As a result, when Gangnam is deformed to the pose that the region is visible to the camera view, as shown in Fig. \ref{fig:failure}, this region shows noise and discontinuity.

\begin{figure}
     \centering
     \begin{subfigure}[b]{0.2\textwidth}
         \centering
         \includegraphics[width=\textwidth]{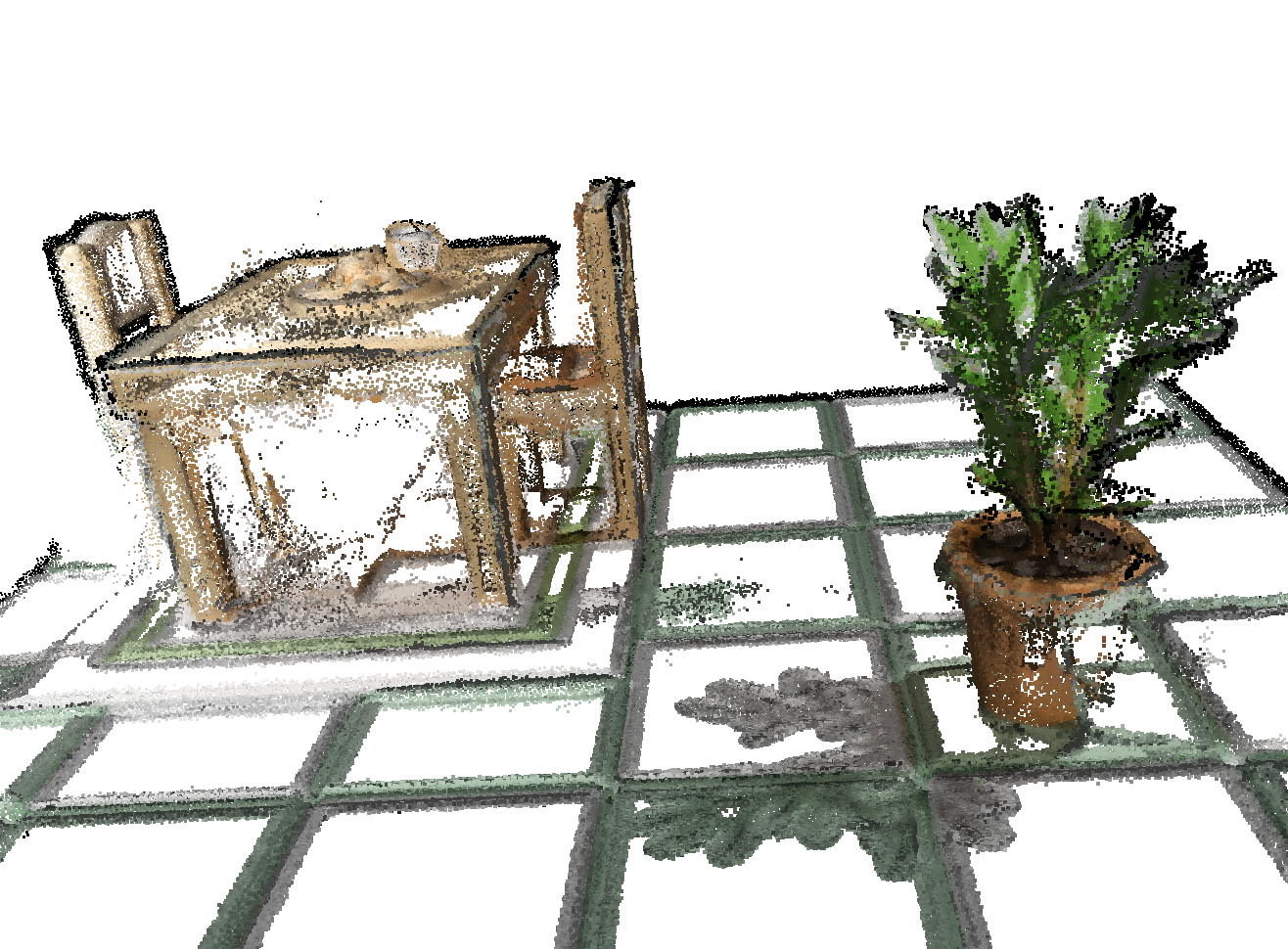}
         \caption{COLMAP}
         \label{fig:bg_colmap}
     \end{subfigure}
     \hfill
     \begin{subfigure}[b]{0.2\textwidth}
         \centering
         \includegraphics[width=\textwidth]{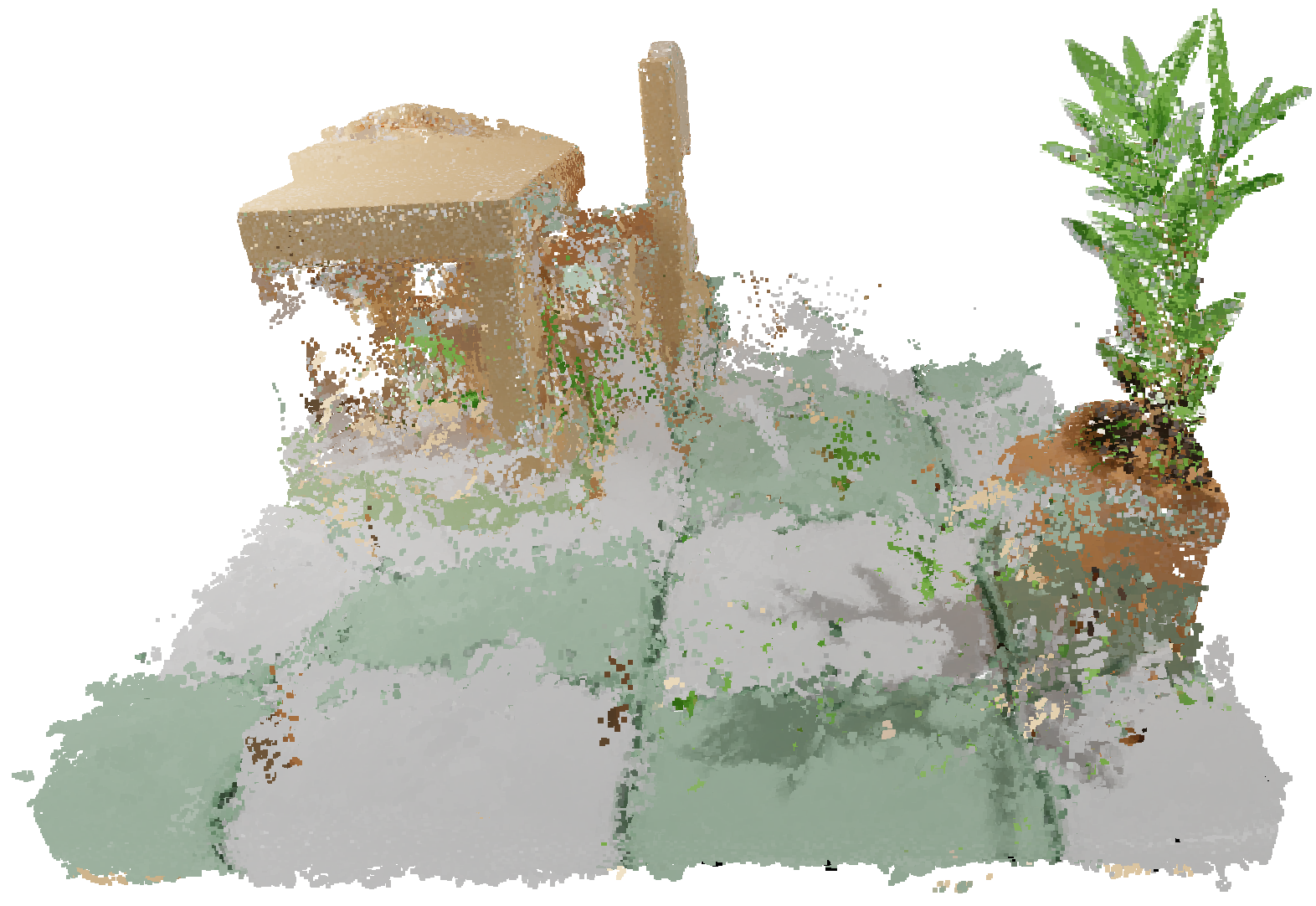}
         \caption{MVSNet}
         \label{fig:bg_mvs}
     \end{subfigure}
     \hfill
     \begin{subfigure}[b]{0.2\textwidth}
         \centering
         \includegraphics[width=\textwidth]{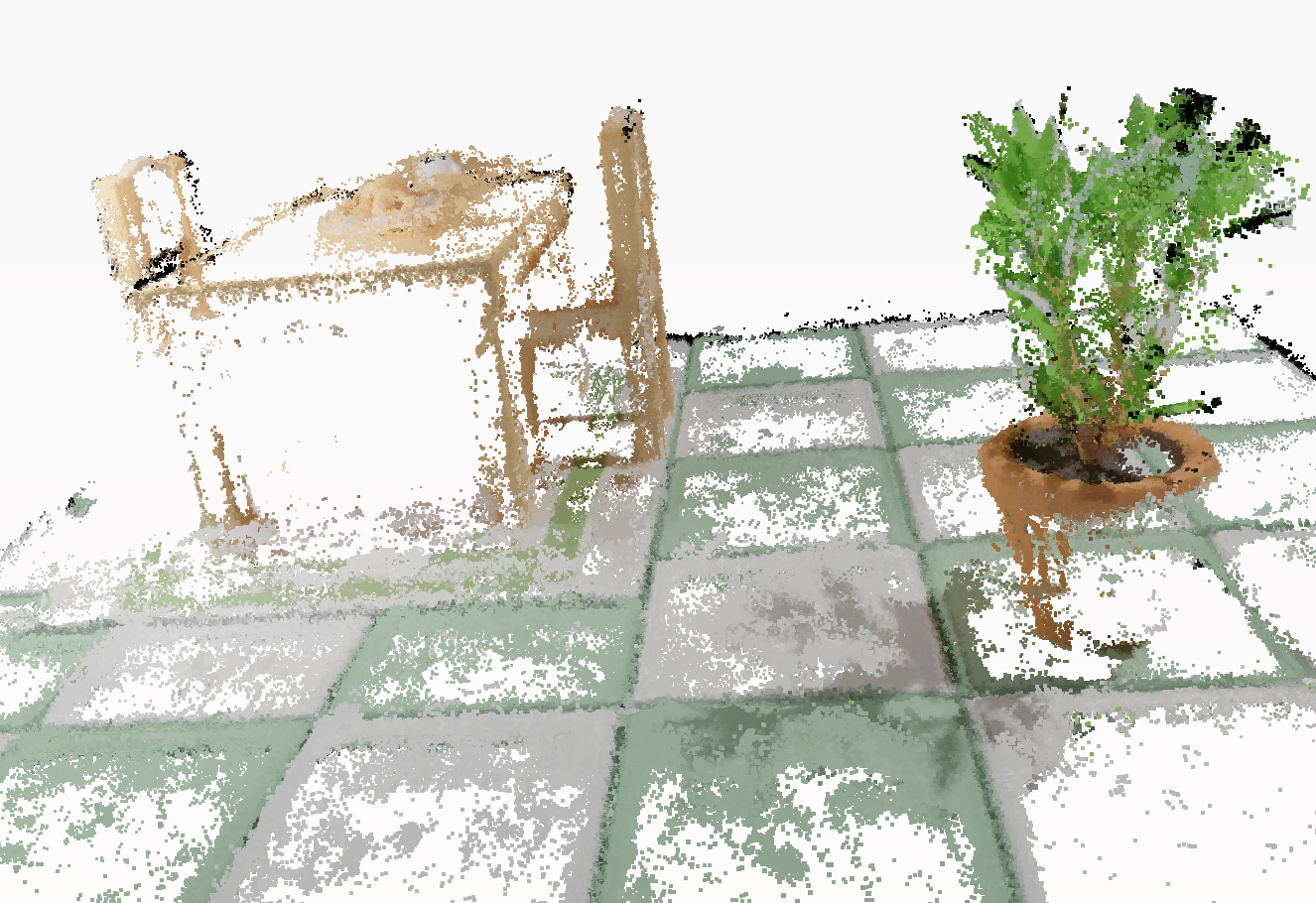}
         \caption{TransMVS}
         \label{fig:bg_transmvs}
     \end{subfigure}
     \hfill
     \begin{subfigure}[b]{0.2\textwidth}
         \centering
         \includegraphics[width=\textwidth]{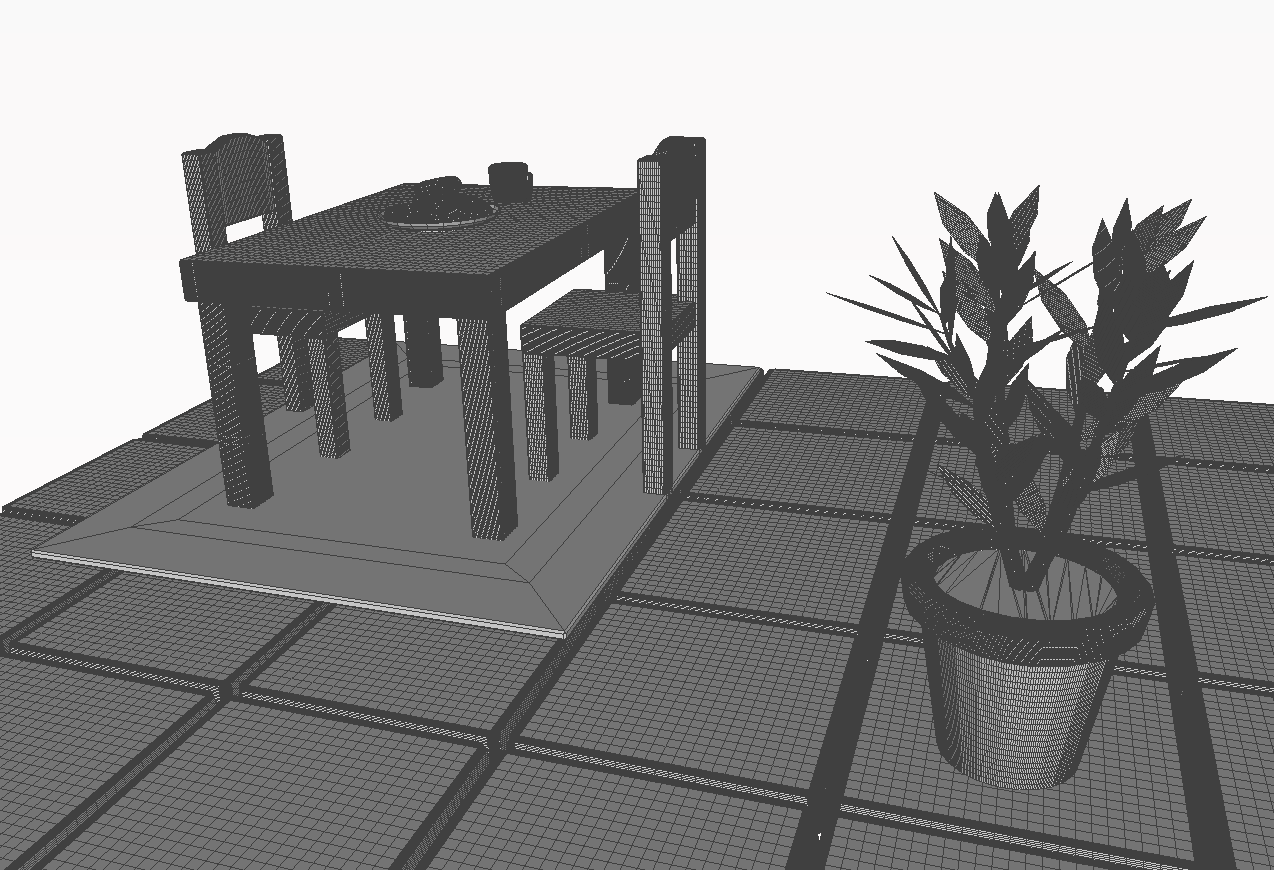}
         \caption{Points from mesh}
         \label{fig:bg_blender}
     \end{subfigure}
        \caption{Point initialization comparison. Among them, both MVSNet and TransMVS were pre-trained on the DTU dataset, and TransMVS was then finetuned on the BlendedMVS dataset. The point cloud extracted from the mesh was directly generated by Blender. }
        \label{fig:background}
\end{figure}

\begin{figure*}[h]
    \centering
        \begin{tabular}{p{16mm}p{20mm}|p{16mm}p{16mm}p{16mm}p{16mm}p{16mm}p{16mm}p{16mm}}
             \includegraphics[width=20mm]{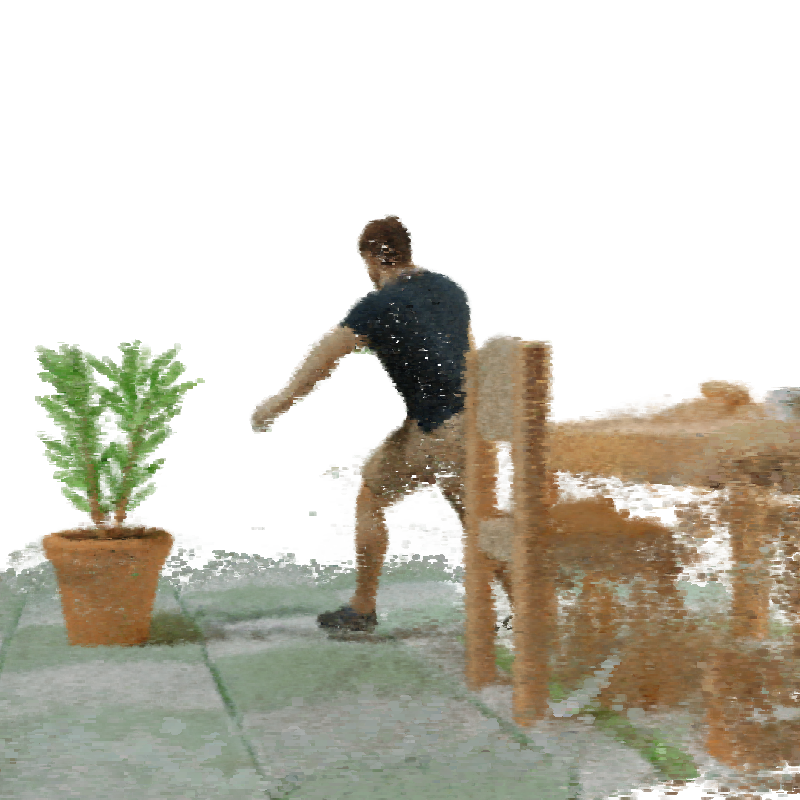}
            & \includegraphics[width=20mm]{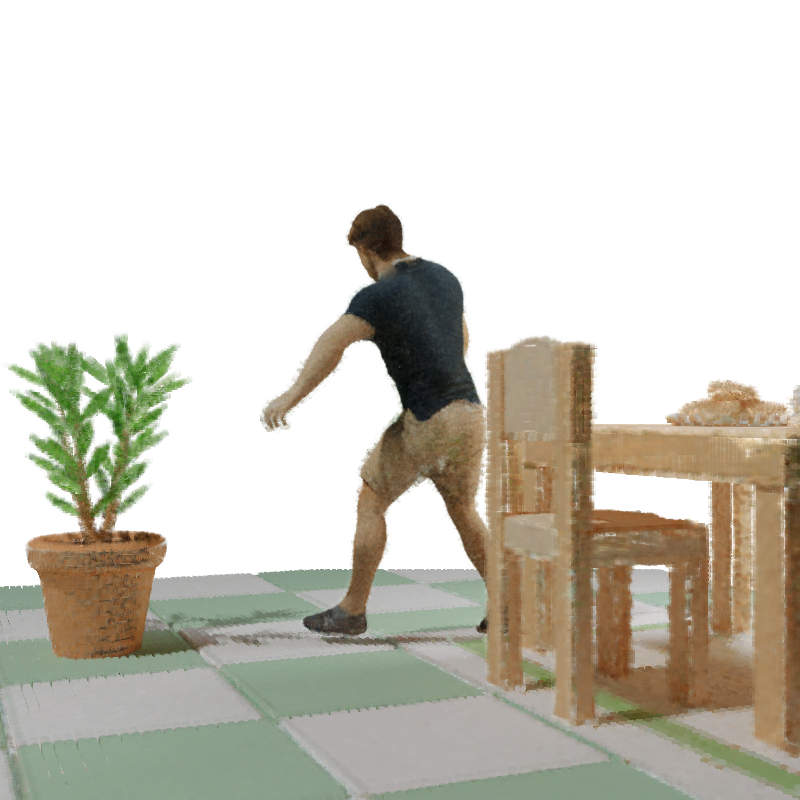}
            & \includegraphics[width=20mm]{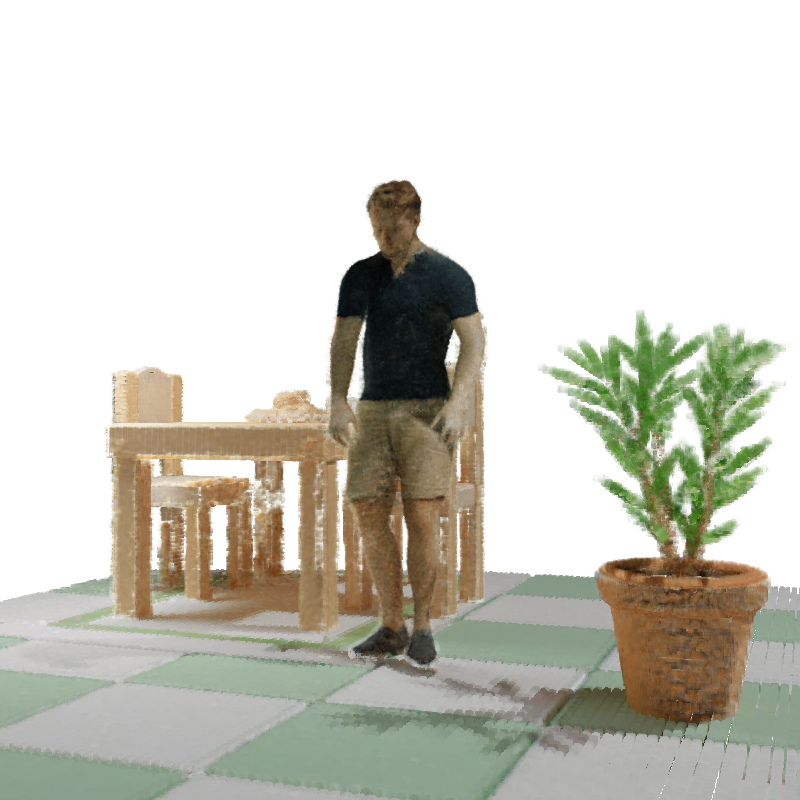}
            & \includegraphics[width=20mm]{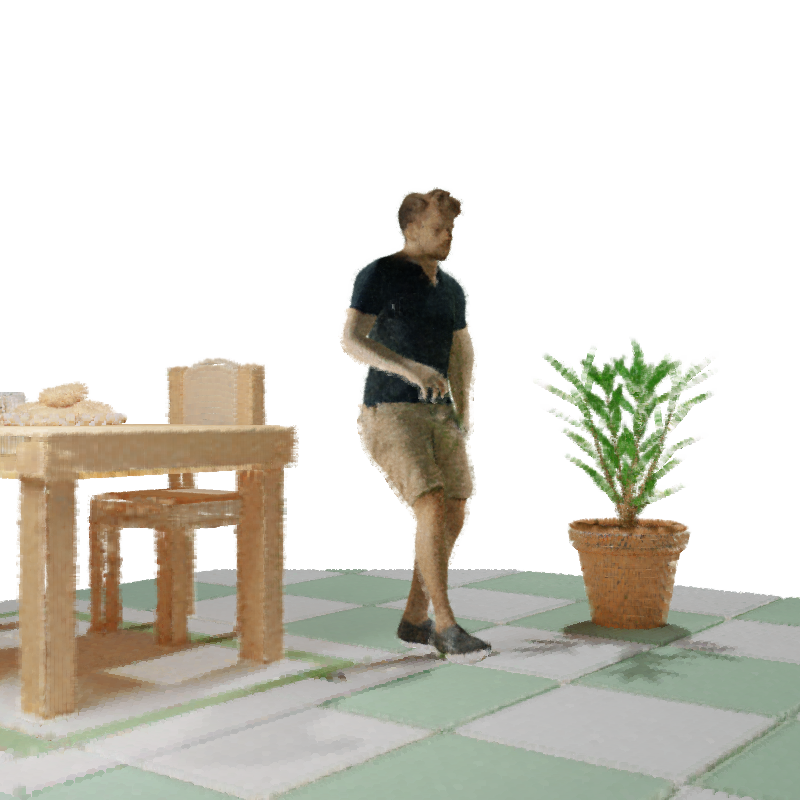}
            & \includegraphics[width=20mm]{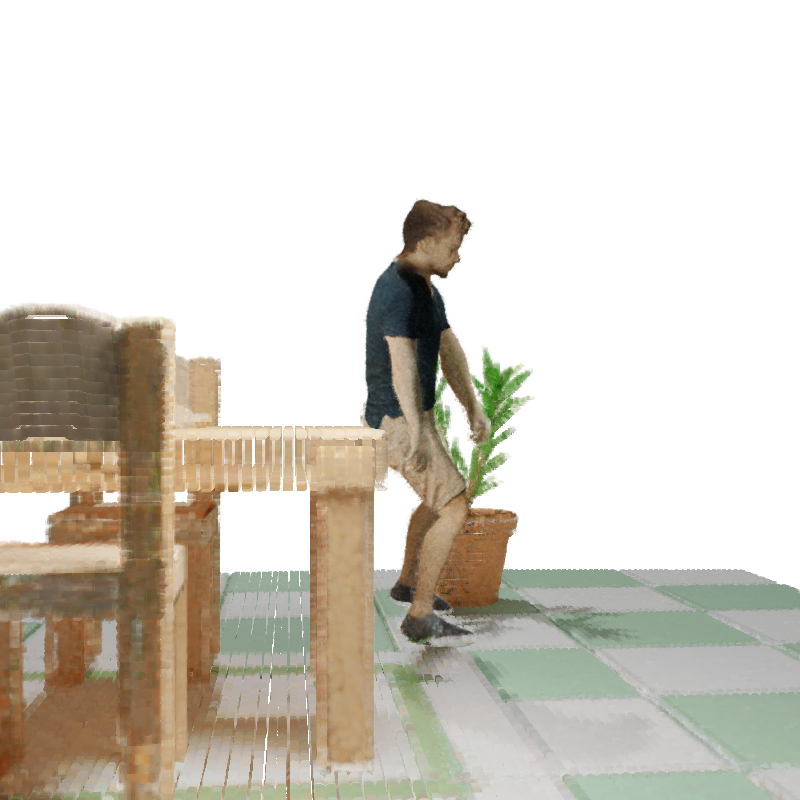}
            & \includegraphics[width=20mm]{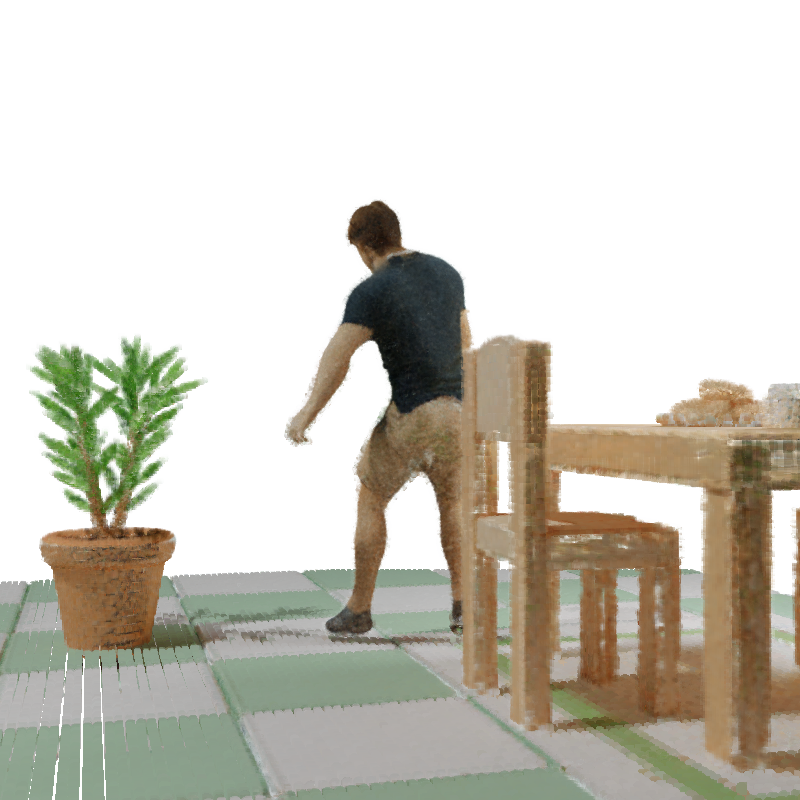}
            & \includegraphics[width=20mm]{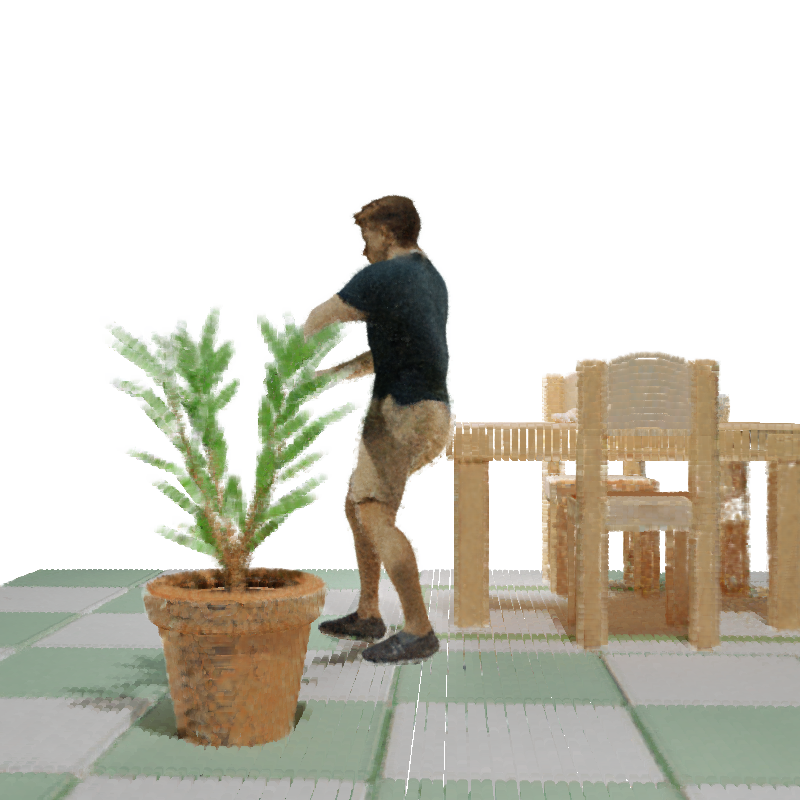}
            & \includegraphics[width=20mm]{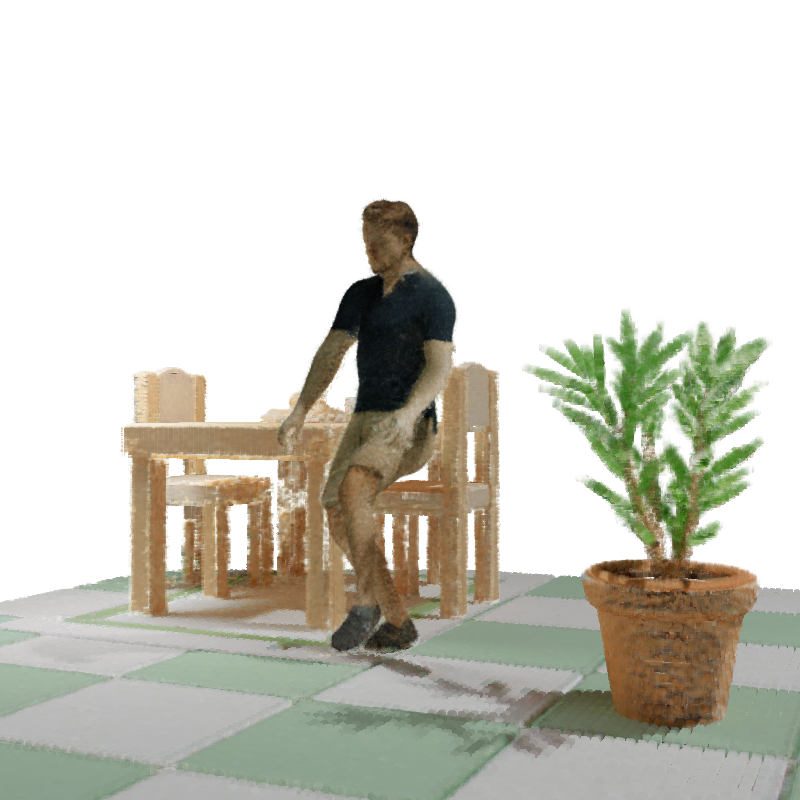}\\
        \end{tabular}%
    \caption{Experiment results of human motion synthesis with background. The left two images of each column represent the multi-view images of the static human initialized with MVSNet and points extracted from the mesh respectively, and the right columns are the novel views of unseen motions.}
    \label{fig:bg_result}
\end{figure*}

\textbf{With Background.} 
Our method can be easily applied to scenes with background, allowing a wider range of applications. However, learning the radiance information of each point highly depends on the initialization of the point-based radiance field. As shown in Fig. \ref{fig:background}, most of the off-the-shelf MVS models fail to provide an initialization precise enough for this process. So instead, we experimented with an initial point cloud extracted from the mesh of the objects in the background, deforming the characters of interest only and combining them in the final rendering result. Fig. \ref{fig:bg_result} illustrates the rendering results of different initialization, as well as the results of deformation together with background.

\section{Conclusion}

In this work, we extended the existing point-based radiance field to support synthesizing controllable human motions, by incorporating it with a surface-based deformation model and a ray-bending procedure. Experimental results show that our method is capable of generating visually faithful dynamic content of both human and nonhuman characters, and can be further extended to scenes with background.

Currently, the reconstruction part of our pipeline is consistent with PointNeRF, which cannot provide the initialization precise enough for satisfactory rendering quality and limits the application of our method. And all the experiments are conducted only on manually synthesized datasets. Future work includes integrating a more powerful MVS model into the pipeline and evaluating our method on the real-world dataset.

\section{Contributions of team members}
\begin{itemize}
    \item Deheng Zhang: Set up the environment. Developed dataset. Integrated PointNeRF and DPF for deformation. Derived the ray bending method and implemented the ray bending code. Run experiments.

    \item Tianyi Zhang: Configured PointNeRF environment. Set up the deformation pipeline and implement the deformation code. Generated datasets. Run PointNeRF and Deformation experiments.

    \item Haitao Yu: Develop dataset. Pipeline design. Derived the ray bending method and implemented the ray bending code. Run experiments.

    \item Peiyuan Xie: Generate dataset. Run experiments. Explored the possibility of performing deformation on scenes with background.

\end{itemize}


{\small
\bibliographystyle{ieee_fullname}
\bibliography{egbib}
}

\end{document}